%% file: emnlp2022.tex
\pdfoutput=1

\documentclass[11pt]{article}

\usepackage[]{EMNLP2022}

\usepackage{times}
\usepackage{latexsym}

\usepackage[T1]{fontenc}

\usepackage[utf8]{inputenc}

\usepackage{microtype}
\usepackage{inconsolata}
\usepackage{amsmath}
\usepackage{amsfonts}
\usepackage{amssymb}
\usepackage{cleveref}
\usepackage{booktabs}
\usepackage{graphicx}
\usepackage{subfigure}
\usepackage{multirow}
\usepackage{amsbsy}
\usepackage{times}
\usepackage{latexsym}
\usepackage{amsfonts}
\usepackage{dsfont}
\usepackage{pgfplots}
\usepackage{diagbox}
\usepackage{xspace}
\usepackage{multirow}
\usepackage{color}
\usepackage{framed}
\usepackage{wrapfig}
\usepackage{tikz}
\usepackage{hyperref}
\usetikzlibrary{intersections}
\input{math_commands.tex}

\usepackage{enumitem}
\usepackage[font=small,labelfont=bf]{caption}
\usepackage{float}

\usepackage{todonotes}

\definecolor{carl}{RGB}{237, 19, 186}

\definecolor{zptu}{RGB}{18, 141, 21}

\definecolor{wx}{RGB}{18, 21, 141}

\newcommand{\MLLM}{{\rm LM}}

\newcommand{\A}{{\rm A}}

\title{
Adapters for Enhanced Modeling of Multilingual Knowledge and Text
}

\author{\normalsize{Yifan Hou$^{1}$, Wenxiang Jiao$^{2}$, Meizhen Liu$^{3}$, Carl Allen$^{1}$, Zhaopeng Tu$^{2}$, Mrinmaya Sachan$^{1}$} \\
  $^{1}$ETH Z\"{u}rich, $^{2}$Tencent AI Lab, $^{3}$Shandong University \\
  $^{1}$\texttt{\{yifan.hou, carl.allen, mrinmaya.sachan\}@inf.ethz.ch} \\
  $^{2}$\texttt{\{joelwxjiao, zptu\}@tencent.com}, $^{3}$\texttt{meizhen.liu@mail.sdu.edu.cn}
}
\begin{document}
\maketitle

\input{./main_text/0abstract.tex}
\input{./main_text/1intro.tex}

\input{./main_text/2problem.tex}
\input{./main_text/3method.tex}

\input{./main_text/4exp.tex}

\input{./main_text/5conclusion.tex}

\bibliography{anthology.bib,custom.bib}
\bibliographystyle{acl_natbib.bst}

\clearpage
\appendix
\input{./main_text/6appendix.tex}

\end{document}

%% file: math_commands.tex
\usepackage{amsmath,amsfonts,bm}

\def\eqref#1{equation~\ref{#1}}

\def\1{\bm{1}}

\def\vb{{\bm{b}}}

\def\vh{{\bm{h}}}

\def\vx{{\bm{x}}}

\def\mK{{\bm{K}}}

\def\mQ{{\bm{Q}}}

\def\mV{{\bm{V}}}
\def\mW{{\bm{W}}}

\DeclareMathAlphabet{\mathsfit}{\encodingdefault}{\sfdefault}{m}{sl}
\SetMathAlphabet{\mathsfit}{bold}{\encodingdefault}{\sfdefault}{bx}{n}



%% file: main_text/0abstract.tex
\begin{abstract}
Large language models appear to learn facts from the large text corpora they are trained on. Such facts are encoded implicitly within their many parameters, making it difficult to verify or manipulate what knowledge has been learned.
Language models have recently been extended to multilingual language models (MLLMs), enabling knowledge to be learned across hundreds of languages. Meanwhile, knowledge graphs contain facts in an explicit \textit{triple} format, which require careful and costly curation and are only available in a few high-resource languages, restricting their research and application. 
To address these issues, we propose to enhance MLLMs with knowledge from multilingual knowledge graphs (MLKGs) so as to tackle language and knowledge graph tasks across many languages, including low-resource ones. Specifically, we introduce a lightweight \textit{adapter set} to enhance MLLMs with cross-lingual entity alignment and facts from MLKGs for many languages.
Experiments on common benchmarks show that such enhancement benefits both MLLMs and MLKGs, achieving: 
(1) comparable or improved performance for knowledge graph completion and entity alignment relative to baselines, especially for low-resource languages (for which knowledge graphs are unavailable); and 
(2) improved MLLM performance on language understanding tasks that require multilingual factual knowledge; 
all while maintaining performance on other general language tasks.\footnote{Our code, models, and data (e.g., integration corpus and extended datasets) are available at \href{https://github.com/yifan-h/Multilingual_Space}{https://github.com/yifan-h/Multilingual\_Space}.}

\end{abstract}

%% file: main_text/1intro.tex
\section{Introduction}
Knowledge graphs serve as a source of explicit factual information for various NLP tasks.
However, language models~\citep{devlin-etal-2019-bert,gpt_brown20}, which capture \textit{implicit} knowledge from vast text corpora, are already being used in knowledge-intensive tasks.
Recently, language models have been successfully extended to multilingual language models (MLLMs) that integrate information sourced across hundreds of languages~\citep{devlin-etal-2019-bert,xlm_conneau19,conneau-etal-2020-unsupervised}. 
However, as with most neural networks, the information is encoded in a diffused and opaque manner that is difficult to interpret, verify or utilize~\citep{lmkb_review_alKhamissi22}.

Meanwhile, multilingual knowledge graphs (MLKGs) require careful curation of explicit facts and annotation of entities that occur in languages (cross-lingual entity alignment), making knowledge graphs expensive and time-consuming to extend to new languages, restricting knowledge graph research to a few high-resource languages.
Further, open-source MLKGs such as WordNet~\citep{bond-foster-2013-linking} and Wikidata~\citep{wikidata_vrandevcic2014} suffer from \textit{incompleteness} as many true facts (or \textit{triples}) and entity alignments are missing~\citep{mtranse_chen17,chen-etal-2020-multilingual}.

\begin{figure}[t!] 
	\centering
	\includegraphics[width = 0.95\columnwidth]{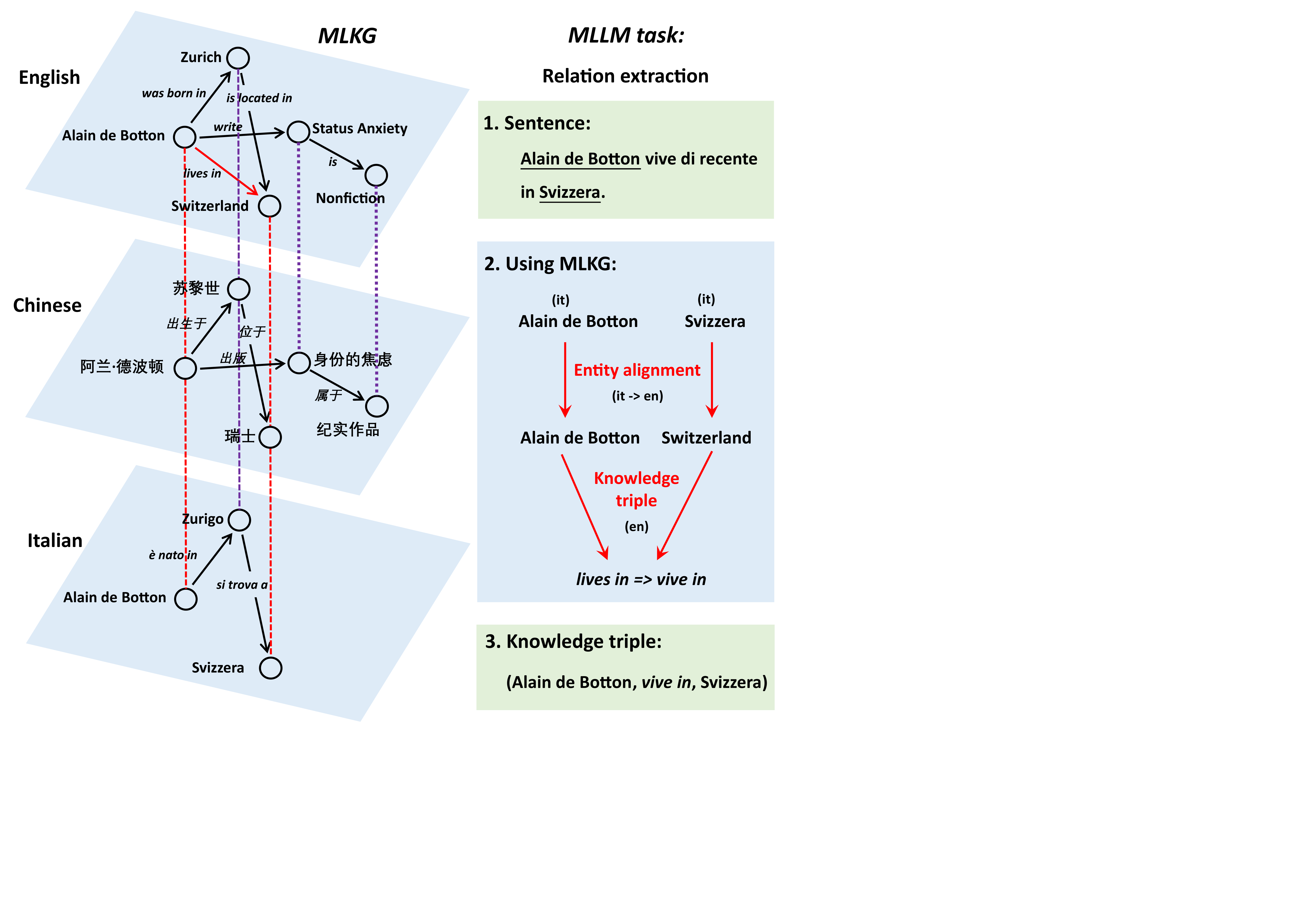}
	\vspace{-.2cm}
	\caption{Combining MLLMs and MLKGs benefits both: MLKGs suffer from incompleteness and are limited to few languages, which MLLMs can supplement. MLLMs lack entity alignment and firm facts, which MLKGs can provide.} 
	\vspace{-.5cm}
	\label{fig:compare}
\end{figure}

In this work, we propose to overcome the above limitations of each knowledge source by integrating MLKGs into MLLMs (as shown in Figure~\ref{fig:compare}), to enable 
(i) the transfer of MLKG knowledge from high-resource languages to low-resource languages; and 
(ii) explicit knowledge of MLKGs to supplement MLLMs for knowledge-intensive language tasks, one of the key challenges in MLLMs~\citep{lmkb_review_alKhamissi22}.

While this idea seems intuitive, there is no easy way to incorporate the explicit knowledge of MLKGs into the parametrically stored information of MLLMs. 
Existing knowledge integration methods utilize language models and knowledge graphs in two ways: 
    (1) training knowledge graph embeddings individually and combining the embeddings corresponding to linked entities in sentences 
    with the language model representations (e.g., KnowBERT~\citep{peters-etal-2019-knowledge} and ERNIE~\citep{zhang-etal-2019-ernie}); or 
    (2) absorbing the knowledge in knowledge graphs into the language model's parameters via joint training (e.g., K-BERT~\citep{kbert_liu19} and K-Adapter~\citep{wang-etal-2021-k}). 

The first method requires embedding knowledge graph entities and accurately extracting entities in sentences across hundreds of languages, which is highly challenging. The second method typically suffers from the \textit{curse of multilinguality}~\citep{conneau-etal-2020-unsupervised,mllmcurse_doddapaneni21,tencent_wmt_jiao22} and \textit{catastrophic forgetting}~\citep{forgetting_kirkpatrick16} due to limited model capacity. 
Most importantly, both methods integrate knowledge implicitly such that it is difficult to access and extend to low-resource languages~\citep{lmkb_review_alKhamissi22}.
%
Furthermore, both methods require large sets of aligned sentences and knowledge triples, which is costly to gather and accurately annotate across hundreds of languages.
%


To address above issues, we first collect and clean multilingual data from Wikidata\footnote{\href{https://www.wikidata.org/wiki/Wikidata:Main_Page}{https://www.wikidata.org/wiki/Wikidata:Main\_Page}} and Wikipedia\footnote{\href{https://en.wikipedia.org/wiki/Main_Page}{https://en.wikipedia.org/wiki/Main\_Page}} for the enhancement, where rich factual knowledge and cross-lingual alignments are available.
Then, we propose to enhance MLLMs with the MLKG information by using a set of \textit{adapters}~\citep{adapter_houlsby19}, which are lightweight, collectively having only around $0.5\%$ extra parameters than the MLLM. 
Each adapter integrates information from either MLKG \textbf{T}riples (i.e.\ facts) or cross-lingual \textbf{E}ntity alignments, and is trained on either \textbf{P}hrase or \textbf{S}entence level data. 
Each of the resulting four adapters (EP/TP/ES/TS) is trained individually to learn information supplemental to that already learned by the MLLM. Adapter outputs are combined by a \textit{fusion} mechanism \citep{pfeiffer-etal-2021-adapterfusion}. Training objectives are similar to those for MLKG embedding~\citep{mtranse_chen17} instead of mask language modeling, which are more efficient with large corpus.

We conduct experiments on various downstream tasks to demonstrate the effectiveness of our approach.
For MLKG tasks, following the data collection methods of two existing benchmarks~\citep{chen-etal-2020-multilingual,mtranse_chen17}, we extended them from 2-5 languages to 22 languages, including two rare languages.\footnote{The extended datasets as well as KI corpus are published with our code implementation.}
Results show that our method obtains comparable performance to existing state-of-the-art baselines on the knowledge graph completion benchmark, and significantly better performance on the entity alignment benchmark. 
More importantly, we can perform these knowledge graph tasks in low-resource languages for which no knowledge graph exists, and achieve comparable results to the high-resource languages. Improvements over baseline MLLMs are significant.
The results demonstrate that our proposed method integrates the explicit knowledge from MLKGs into MLLMs that can be used across many languages.
Our method also improves existing MLLMs noticeably on knowledge-intensive language tasks, such as cross-lingual relation classification, whilst maintaining performance on general language tasks such as named entity recognition (NER) and question answering (QA).

%% file: main_text/2problem.tex
\section{Multilingual Knowledge Integration}

In this paper, we fuse knowledge from a MLKG into a MLLM. Following previous works~\citep{wang-etal-2021-k,mlkidamo_liu21}, we make use of an entity tagged corpus of text (called a \textit{knowledge integration corpus}) for knowledge integration. We formally introduce these concepts below.

\paragraph{MLLM.} A multilingual LM can be thought of as an encoder that can represent text in any language $l$ in a set of languages $\mathcal{L}$. Let
$\mathcal{V}$ denote the shared vocabulary over all languages. Let $t^{l} \!\in\! \mathcal{V}$ denote a token in language $l$. A 
sentence $s^l$ in a language $l$ 
can be denoted as a sequence of tokens: $s^{l} \!=\! ( t_{1}^{l}, t_{2}^{l},... )$.
The output representations of the MLLM for $s^{l}$ can be denoted by a sequence of vectors: $\MLLM(s^{l}) \!=\! ( \vh_{1}, \vh_{2},... )$. These vectors correspond to representations for each token in the sentence, one representation per input token. Various tokenization schemes such as wordpiece or BPE might be considered here.
We use the average of the token representations as the representation of the sentence: $\overline{\MLLM(s^{l})} \!=\! \text{mean}( \vh_{1}, \vh_{2},... )$. Similarly, for a phrase $s_{ij}^{l}$ (starting from the $i$-th token and ending in the $j$-th token in the sentence), we can obtain its contextualized representation as $\overline{\MLLM(s_{ij}^{l})} = \text{mean}( \vh_{i}, \vh_{i+1},\dots \vh_{j})$. 

\paragraph{MLKG.} A multilingual knowledge graph is a graph with entities and knowledge triples in each language $l \in \mathcal{L}$.
Let $\mathcal{E}$ denote the set of entities and $\mathcal{T}$ denote the set of knowledge triples. In a MLKG, each entity indexed $i$ might appear in several languages. Let $e^{l}_{i}$ denote the entity label of the $i$-th entity in language $l$. Furthermore, we denote a knowledge triple in the MLKG as $(e^{l}_{i}, r^{l''}_{k}, {e}^{l'}_{j}) \in \mathcal{T}$, where $r^{l''}_{k}$ is the $k^{th}$ relation. Note that since entities (as well as relations) may appear in various languages under different labels, knowledge triples can be defined across languages.

\paragraph{Knowledge Integration Corpus.} 
For knowledge integration, besides the MLKG, we make use of a corpus of text $\mathcal{C}$ (as shown in the right part of Figure~\ref{fig:adapter}). The corpus $\mathcal{C}$ comprises of two kinds of texts. First, we have a set of texts $\mathcal{C}_1$ for the cross-lingual entity alignment, which comprise of sentences with mentions of entities in the MLKG. For example in Figure~\ref{fig:adapter}, given the sentence \textit{De Botton spent his early years in Zurich}, we have the aligned entity \textit{Zurich} and its cross-lingual labels.
The second set of texts $\mathcal{C}_2$ is for the knowledge triple, which comprises of sentences aligned with knowledge triples in the MLKG. For example in Figure~\ref{fig:adapter}, given the sentence \textit{Zurich is the largest city in Switzerland}, we have its aligned knowledge triple \textit{(Zurich, is located in, Switzerland)}.

%% file: main_text/3method.tex
\begin{figure*}[!htbp] 
	\centering
	\includegraphics[width = 1.8\columnwidth]{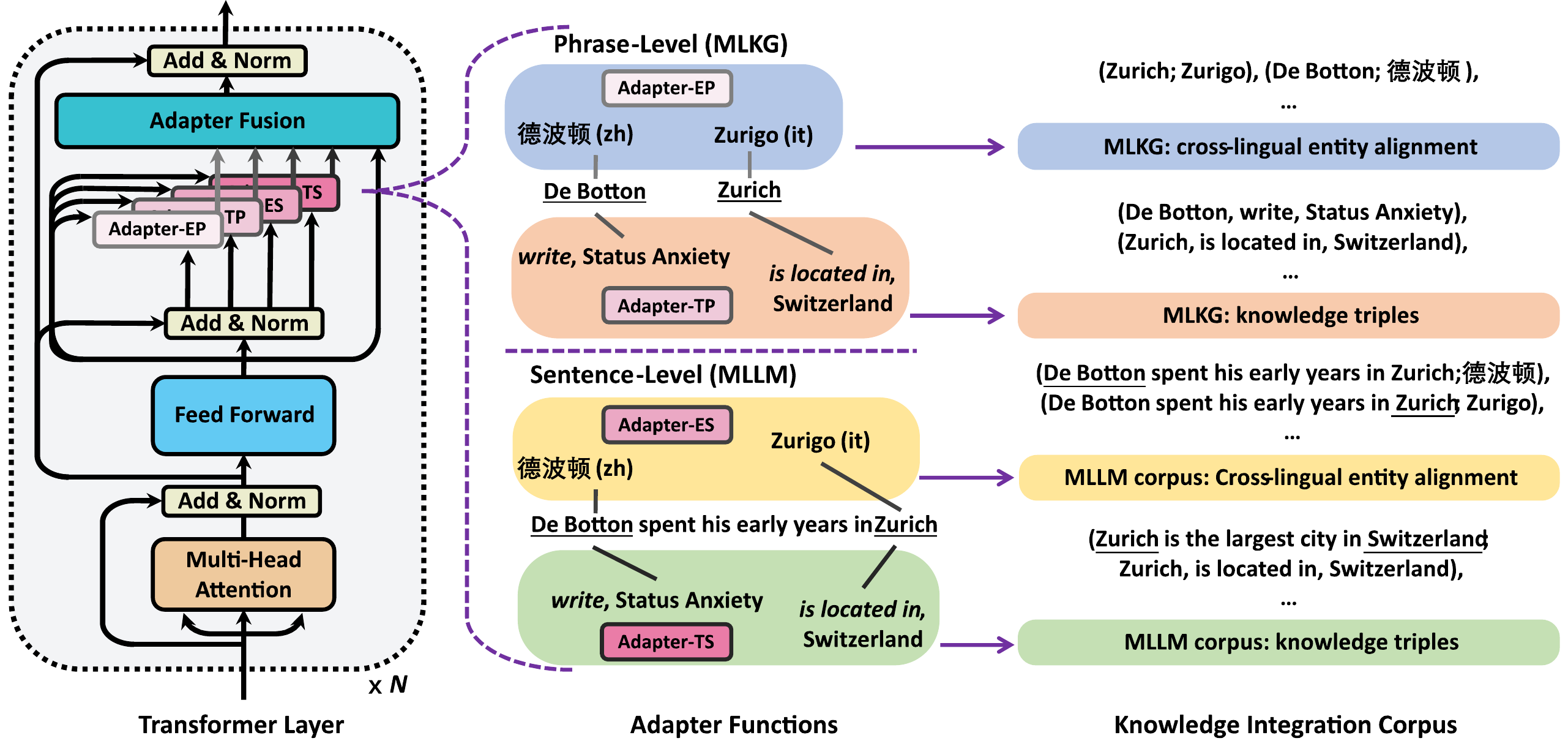}
	\vspace{-.2cm}
	\caption{The architecture of MLLMs with adapters and their roles. We enhance multilingual and factual knowledge in phrase and sentence levels using different knowledge integration corpus.}
	\vspace{-.4cm}
	\label{fig:adapter}
\end{figure*}

\section{Adapters and Adapter Fusion}
In this section, we first describe how we incorporate adapters into language models and how they can be used to enhance them with different sources of knowledge from knowledge graphs.

\paragraph{Adapter.}
Adapters have become a popular choice for parameter-efficient finetuning of language models on downstream tasks~\citep{adapter_houlsby19} due to their flexibility, effectiveness, low cost and scalability~\citep{pfeiffer-etal-2021-adapterfusion}. 
Adapters are new modules that are added between layers of language models\footnote{Where to insert adapters is flexible but a common choice is after the feedforward layer of a transformer layer.}, the parameters of which are updated only during finetuning while the language model parameters are frozen.
An adapter is a bottleneck layer composed of two feed-forward layers with one non-linear activation function. 
For $\vh^{m}$, the hidden representation of token $t^{l}_i$ at layer $m$, the adapter acts as
\begin{equation}\label{eq:adapter}\small
    \A (\vh^{m}) = \mW_{\text{up}} \cdot \sigma (\mW_{\text{down}} \cdot \vh^{m} + \vb_{\text{down}}) + \vb_{\text{up}}.
\end{equation}
Here, $\mW_{\text{down}}$ and $\mW_{\text{up}}$ are weight matrices, which map the hidden representations to the low-dimensional space and then map them back.
$\vb_{\text{down}}$ and $\vb_{\text{up}}$ are bias parameters, and $\sigma$ is a nonlinear activation function.

\paragraph{Adapter Fusion.}
We follow the architecture of~\citet{pfeiffer-etal-2021-adapterfusion}, but instead of using adapters for finetuning, we use them to enhance MLLMs with knowledge. Our approach is similar to~\citet{wang-etal-2021-k}, but our adapters supplement and augment the existing implicit knowledge of MLLMs (into the explicit geometric properties of hidden representations), 
And our approach is more lightweight, with only c.$0.5\%$ additional parameters (\textit{cf} $>\!10\%$ in \citet{wang-etal-2021-k}).

As shown in Figure~\ref{fig:adapter} (left), still considering the $m$-th layer, the output representations of the feedforward layer (denoted $\vh^{m}$ as in Eq.~\ref{eq:adapter}) are input to the adapters. A fusion layer aggregates all adapter outputs $\A_n(\vh^{m})$ ($n \!\in\! \{1...N\}$ indexes each adapter) and the un-adapted representations with a multiplicative attention mechanism:

\vspace{-14pt}
\begin{equation}\label{eq:fusion}\notag\small
\begin{aligned}
    \A_{\text{fusion}}(\vh^{m}) &= \sum_{n=0}^{N} a_n^{m} \cdot \mV^{m} \cdot \A_n(\vh^{m}), 
    \\
    a_n^{m} & = \text{softmax}(\vh^{m}\mQ^{m} \otimes \A_n(\vh^{m})\mK^{m}).
\end{aligned}
\end{equation}
Here, $\A_0(\cdot)$ is the identity function; $\mQ^{m}$, $\mK^{m}$, $\mV^{m}$ are parameters in the multiplicative attention mechanism; and $\otimes$ is the Hadamard product.

The additional knowledge to be learned by the adapters comes from knowledge \textbf{T}riples and \textbf{E}ntity alignments, each provided in both \textbf{P}hrase and \textbf{S}entence format (hence $N\!=\!2\times2\!=\!4$).
As shown in Figure~\ref{fig:adapter} (center), for a given entity in two languages $l$ and $l'$, {\colorbox[rgb]{0.706,0.780,0.906}{{Adapter-EP.}}} learns to align the two (multilingual) representations of ${e}^{l}_{i}$ and ${e}^{l'}_{i}$, e.g., \textit{Zurich} is aligned with \textit{Zurigo}. {\colorbox[rgb]{0.973,0.796,0.678}{{Adapter-TP.}}} learns knowledge triples, e.g., predicting \textit{Switzerland} given entity and relation \textit{(Zurich, is located in,)}. 
Besides these \textit{non-contextualized} settings, entities within context can be considered also (MLLM corpus). Thus, {\colorbox[rgb]{1.,0.902,0.6}{{Adapter-ES.}}} and {\colorbox[rgb]{0.773,0.878,0.76}{{Adapter-TS.}}} have the similar objectives but use contextualized representations from input sentences.

\section{Knowledgeable Adapters}
Next, we design objectives with corresponding knowledge integration datasets to train a set of adapters. Similar to MLKG embedding~\citep{mtranse_chen17}, we aim to encode knowledge into the geometric properties of the \textit{adapted} MLLM representations, i.e., the MLLM and adapters collectively act as an MLKG embedding model. 
Specifically, we use cosine distance within the contrastive learning loss of InfoNCE~\citep{infonce_oord18}:
\begin{equation}\notag\small
\begin{aligned}
    {\text{INCE}}(\vx,\vx') & =  \log \frac{\cos(\vx, \vx')}{\sum_{\vx'' \in X} \cos(\vx, \vx'')} ,
\end{aligned}
\end{equation}
where $X$ is a batch that includes the positive sample $\vx'$ and a number of negative samples.\footnote{We use \textit{in-batch} negative sampling, where entities (with labels in any languages) in the batch are randomly selected.}

\paragraph{{\colorbox[rgb]{0.706,0.780,0.906}{{Adapter-EP.}}}}
We use Wikidata~\citep{wikidata_vrandevcic2014} to enhance MLLMs with the knowledge of cross-lingual entity alignments. Inspired by the idea that languages are aligned implicitly in a universal space in MLLMs~\citep{wu-dredze-2019-beto,universalspace_wei21}, we train the aligned entities to have closer representations. 
Denoting the MLLM with this adapter as $\MLLM(\cdot)$, the objective used to train EP is:
\begin{equation}\notag\small
\begin{aligned}
     \mathcal{L}_{\text{EP}} = \sum_{({e}^{l}_{i}, {e}^{l'}_{i}) \in \mathcal{E}} {\text{INCE}} {\left(\overline{\MLLM({e}^{l}_{i})}, \overline{\MLLM({e}^{l'}_{i})}\right)},
\end{aligned}
\end{equation}
where $\overline{\MLLM(\cdot)}$ means we take the mean of token representations as the entity representation vector. 

\paragraph{{\colorbox[rgb]{0.973,0.796,0.678}{{Adapter-TP.}}}}
We train this adapter using the knowledge triples in Wikidata. Inspired by previous knowledge graph embedding algorithms~\citep[e.g.][]{transe_bordes13}, for a given fact triple, we train the (adapted) object entity embedding to be close to the (adapted) joint embedding of the subject entity and relation. The objective used to train TP is quite different from existing mask language modeling-based ones:
\begin{equation}\notag\small
\begin{aligned}
     \mathcal{L}_{\text{TP}} = \!\!\!\!\!\!\!\! \sum_{({e}^{l}_{i}, r^{l''}_k, {e}^{l'}_{j}) \in \mathcal{T}} \!\!\!\!\!\!\!\! {\text{INCE}}{\left(\overline{\MLLM([{e}^{l}_{i};r^{l''}_k])}, \overline{\MLLM({e}^{l'}_{j})}\right)},
\end{aligned}
\end{equation}
where $[;]$ denotes text concatenation. Note that we apply \textit{code-switching} \citep{mlkidamo_liu21}, and thus entities and relations can be in different languages. This is helpful to capture knowledge triples for low-resource languages.

\begin{figure*}[!thbp]
	\centering
	\includegraphics[scale=.15]{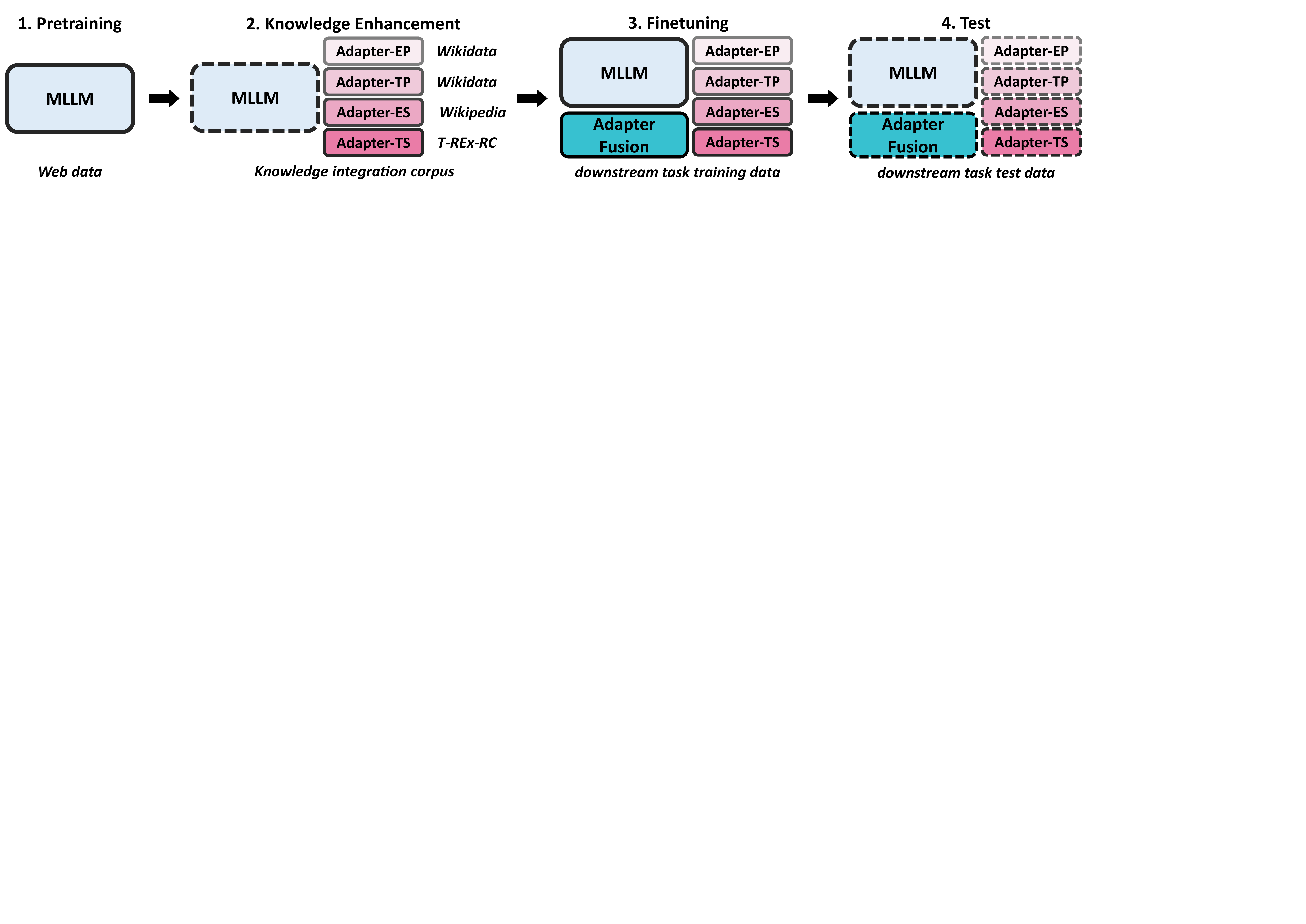}
	\vspace{-.2cm}
	\caption{Four stages of using the knowledge adapter set in MLLMs. The dashed outlines mean the parameters are frozen.}
	\vspace{-.4cm}
	\label{fig:pattern}
\end{figure*}

\paragraph{{\colorbox[rgb]{1.,0.902,0.6}{{Adapter-ES.}}}}
Entity alignment can also be applied to contextualized embeddings produced by the MLLM when entities are input within natural language sentences. For this purpose, we use summaries taken from multilingual Wikipedia. Specifically, we first align the entity in Wikidata with the Wikipedia title, and extract sentences that contain the entity label in its summary. As described earlier, we denoted this corpus as $\mathcal{C}_1$.
Thus, similar to {Adapter-EP}, we train ES by aligning contextualized entity representations of cross-lingually aligned entities with the objective:
\begin{equation}\notag\small
\begin{aligned}
     \mathcal{L}_{\text{ES}} = \sum_{({e}^{l'},{s}^{l}) \in \mathcal{C}_1} {\text{INCE}}{\left(\overline{\MLLM({{s}^{l}_{ij}}) }, \overline{\MLLM({e}^{l'})}\right)},
\end{aligned}
\end{equation}
where ${s}^{l}_{ij}$ means that we input sentence ${s}^{l}$ into an MLLM but keep only the representation of entity label ${e}^{l}$ (indexed from $i$-th token to $j$-th token). 
As in Figure~\ref{fig:adapter} (right), ${s}^{l}$ is: \textit{De Botton spent his early years in Zurich}, and ${s}^{l}_{ij}$ here is the entity label of ${e}^{l}$ as: \textit{Zurich}. 
The difference between this adapter and {Adapter-EP} is that contextual information is included within the entity representation.

\paragraph{{\colorbox[rgb]{0.773,0.878,0.76}{{Adapter-TS.}}}}
Knowledge triples can also be learned with contextualized embeddings. This requires paired data in which triples (entities and relations) are annotated in natural sentences. However, no such multilingual corpus exists. Thus, we use the T-REx-RC dataset~\citep{elsahar-etal-2018-rex}\footnote{We denoted this aligned corpus earlier by $\mathcal{C}_2$}, which provides aligned data in English and contains sentence and triple pairs. 
Thus, the objective used to train TS is:
\begin{equation}\notag\small
\begin{aligned}
     \mathcal{L}_{\text{TS}} = \!\!\!\!\!\!\!\! \sum_{(s_k, (e_i, r, e_j)) \in \mathcal{C}_2} \!\!\!\!\!\!\!\! {\text{INCE}}{\left(\overline{\MLLM(s_k \backslash e_j)}, \overline{\MLLM({e}_{j})}\right)},
\end{aligned}
\end{equation}
where $s_k \backslash e_j$ represents the sentence $s_k$ with entity label $e_j$ masked. 
As the example in Figure~\ref{fig:adapter} (right), $s_k \backslash e_j$ is: \textit{[MASK] is the largest city in Switzerland}, and the aligned triple is: \textit{(Zurich, is located in, Switzerland}. 
In contrast to {Adapter-TP}, subject entities and relations occur in natural sentences.

\subsection{Enhancement Workflow}
We introduce our overall enhancement workflow, which contains four stages.
In the first stage, an MLLM is \textit{pretrained} on a large amount of data. In the second stage, the MLLM is frozen while each adapter is trained separately on its particular dataset (knowledge integration corpus) to extract additional information. Adapter outputs are aggregated in the fusion layer to enable their collective knowledge to be pooled~\citep{pfeiffer-etal-2021-adapterfusion}. 
For example, we lack knowledge graph data for low-resource languages, however we have two adapters (TP, TS) that learn facts in a particular language (English) and two adapters (EP, ES) that learn cross-lingual alignment. By aggregating them, we can effectively integrate factual knowledge into the representations of low-resource languages. 
In the third and final stages, all parameters of the MLLM, the adapters, and the fusion module are \textit{finetuned} on a training set for a specific downstream task resulting in a specialized model for the task (see Figure~\ref{fig:pattern}).


%

%% file: main_text/4exp.tex
\section{Experiments}
This section first introduces the general experimental settings~(\S\ref{exp:setting}).
We then show that our adapter set can enhance MLLMs with the knowledge of MLKGs and, in particular, that the enhanced MLLMs generalize well to perform MLKG-related tasks in low-resource languages~(\S\ref{exp:mlkg}).
We also show that enhancing MLLMs with MLKGs improves their performance on knowledge-intensive language tasks~(\S\ref{exp:mllm}). We compare our approach with the only existing MLKG integration work~(\S\ref{exp:compare}).
Finally, we present an ablation study of the adapter set to demonstrate the effectiveness of each adapter~(\S\ref{exp:ablation}).

\subsection{MLLMs and Integration Corpus}\label{exp:setting}
We select three representative MLLMs implemented by Huggingface\footnote{\href{https://huggingface.co/}{https://huggingface.co/}} and train a set of adapters for each: 
the base version of mBERT~\citep{devlin-etal-2019-bert}, 
and both the base and large versions of XLMR (i.e., XLM-RoBERTa)~\citep{conneau-etal-2020-unsupervised}.
Since mBERT and XLMR cover different sets of languages, we consider the intersecting $84$ languages supported by both models. 
All adapters are trained with the same hyperparameters (see Appendix~\ref{appendix:implementation} for details).

\begin{table}[!htbp]
    \centering
    \vspace{-.1cm}
    \caption{
    Statistics of knowledge integration corpora for training adapters. \textit{Align.}: all aligned multilingual entities; \textit{Relat.}: all relations in triples; \textit{Sent.}: sentences.}
    \vspace{-.2cm}
    \resizebox{1.\columnwidth}{!}{
        \begin{tabular}{c|c|l}
        \toprule
        \bf Module & \bf Source & \qquad\qquad \bf Statistics \\
        \midrule
        Adapter-EP & Wikidata (MLKG) & Entity / Align.: 1.55M / 63.25M \\
        Adapter-TP & Wikidata (MLKG) & Triple / Relat.: 9.42M / 1422 \\
        Adapter-ES & Wikipedia ($\mathcal{C}_1$) & Entity / Sent.: 0.20M / 1.93M \\
        Adapter-TS & T-REx-RC ($\mathcal{C}_2$) & Triple-Sent. Pair: 0.97M \\
        \bottomrule
        \end{tabular}
    }
    \vspace{-.2cm}
    \label{tab:ki_statistics}
\end{table}
The statistics of the knowledge integration corpora are summarized in Table~\ref{tab:ki_statistics}.
Next, we introduce their preprocessing steps.
The set of entity alignments used to train \textbf{Adapter-EP} is extracted from Wikidata by keeping only entities that have more than $10$ multilingual entity labels among the $84$ considered languages. Knowledge graph triples are used to train \textbf{Adapter-TP} if both entities are in that entity set (see Table~\ref{tab:distribution-wikidata} of Appendix~\ref{appendix:ki_dataset} for further details). 
%
For the Wikipedia dataset, we use entities in the Wikidata subset and query their descriptions (the first sentence in the Wikipedia summary that contains the entity label). 
We remove entities that have less than $2$ multilingual descriptions, which results in 1.93 million multilingual sentences to train \textbf{Adapter-ES}.
For {\textbf{Adapter-TS}}, we use the monolingual dataset T-REx-RC~\citep{elsahar-etal-2018-rex}, which has 0.97 million alignments between knowledge triples and sentences in English.

\subsection{MLKG Benchmarks}\label{exp:mlkg}
We show that \textit{our knowledge adapter set can enhance MLLM performance at MLKG-related tasks}.
We select two popular MLKG benchmarks for evaluation: DBP5L~\citep{chen-etal-2020-multilingual} for the knowledge graph completion task, and WK3L~\citep{mtranse_chen17} for the cross-lingual entity alignment task. These tasks require the MLLM to identify the correct entity, which is performed by maximizing the similarity of output representations.

\begin{figure}[!htbp]
	\centering
	\vspace{-.4cm}
	\subfigure{
	    \pgfplotsset{width=7.8cm, height=4.2cm}
		\begin{tikzpicture}[font=\tiny]
		\begin{axis}[ybar, bar width=0.14cm, enlarge x limits=.1, legend columns=-1,
                    symbolic x coords={el, en, es, fr, ja, ast, ca, da, de, fa, fi, hu, it, nb, nl, pl, pt, ru, sv, zh, eo, vo}, 
                    y label style={at={(axis description cs:0.08,0.5)},anchor=south},
                    x tick label style={rotate=45, anchor=east, align=left},
                    ylabel={\# of knowledge triples (MLKG completion)}, 
                    nodes near coords align={vertical},
                    every node near coord/.append style={font=\fontsize{1}{1}\selectfont},
                    ymin=0, ymax=6500, xtick={el, en, es, fr, ja, ast, ca, da, de, fa, fi, hu, it, nb, nl, pl, pt, ru, sv, zh, eo, vo}]
		\addlegendentry{Sup}
		\addlegendentry{ZS-In}
		\addlegendentry{ZS-Un}
		\addplot coordinates {(el,1082) (en,5984) (es,4101) (fr,4436) (ja,2569)};
		\addplot[xshift=-.18cm,fill=white] coordinates {(ast,2823) (ca,2959) (da,2566) (de,4059) (fa,2329) (fi,2582) (hu,2558) (it,3614) (nb,2717) (nl,4316) (pl,2998) (pt,3184) (ru,2887) (sv,2993) (zh,2591)};
		\addplot[xshift=-.36cm,fill=black] coordinates {(eo,963) (vo,164)};
		\end{axis}
		\end{tikzpicture}
	}	
	\subfigure{
	    \pgfplotsset{width=7.8cm, height=4.2cm}
		\begin{tikzpicture}[font=\tiny]
		\begin{axis}[ybar, bar width=0.14cm, enlarge x limits=.1, legend columns=-1,
                    symbolic x coords={en-fr, en-de, en-ar, en-ast, en-ca, en-cs, en-da, en-es, en-fa, en-fi, en-hu, en-it, en-ja, en-nb, en-nl, en-pl, en-pt, en-ru, en-sv, en-zh, en-eo, en-vo}, 
                    y label style={at={(axis description cs:0.08,0.5)},anchor=south},
                    x tick label style={rotate=45, anchor=east, align=left},
                    ylabel={\# of alignments (entity alignment)}, 
                    nodes near coords align={vertical},
                    every node near coord/.append style={font=\fontsize{1}{1}\selectfont},
                    ymin=0, ymax=43000, xtick={en-fr, en-de, en-ar, en-ast, en-ca, en-cs, en-da, en-es, en-fa, en-fi, en-hu, en-it, en-ja, en-nb, en-nl, en-pl, en-pt, en-ru, en-sv, en-zh, en-eo, en-vo}]
		\addlegendentry{Sup}
		\addlegendentry{ZS-In}
		\addlegendentry{ZS-Un}
		\addplot coordinates {(en-fr,39155) (en-de, 41018)};
		\addplot[xshift=-.18cm,fill=white] coordinates {(en-ar,16818) (en-ast,19834) (en-ca,22567) (en-cs,16570) (en-da,20093) (en-es,28288) (en-fa,16120) (en-fi,20608) (en-hu,18896) (en-it,26393) (en-ja,22012) (en-nb,20748) (en-nl,29378) (en-pl,21535) (en-pt,23001) (en-ru,22665) (en-sv,22986) (en-zh,20891) };
		\addplot[xshift=-.36cm,fill=black] coordinates {(en-eo,8913) (en-vo,2954)};
		\end{axis}
		\end{tikzpicture}
	}
	\vspace{-.5cm}
	\caption{Statistics of the size of test sets for MLKG completion and entity alignment tasks. We can see that extended test sets for zero-shot languages have comparable number of samples as original test sets.}
	\vspace{-.2cm}
	\label{fig:extendedkg}
\end{figure}
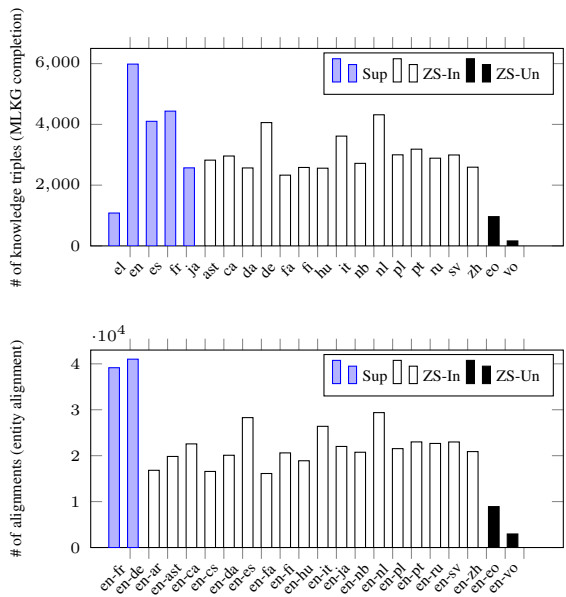

To evaluate MLLMs in a more comprehensive setting, we extend their test sets (from $2\!-\!5$ languages) to $22$ languages following their data construction settings\footnote{We follow settings of data collection pipelines described in~\citet{chen-etal-2020-multilingual,mtranse_chen17} for the extension.}, where languages that contain the most entity labels are selected. Statistics are in Figure~\ref{fig:extendedkg}.
We split these languages into three categories to show the generalizability of enhanced MLLMs: 
\textbf{Sup.}: \textit{supervised languages}, which are used to train adapters and for finetuning; 
\textbf{ZS-In}: \textit{zero-shot languages}, which are used for adapter training but not for finetuning; \textbf{ZS-Un.}: \textit{unseen languages}, which are unseen in both adapter training and finetuning.

\subsubsection{Knowledge Graph Completion}
The knowledge graph completion task tests if the model can find the missing triples in different languages. Specifically, for each test triple of a given language, the model is asked to retrieve the correct object entity from the entity set of that language given the subject entity and relation.

\paragraph{Settings.}
We follow the settings of DBP5L.\footnote{Note that some entity alignments across $5$ languages are provided. We only consider the triple data for simplicity and test entity alignment with another benchmark: WK3L.} Specifically, we use the training set of knowledge triples of the five languages (i.e.\ the Sup.\ set) to finetune the model, and then use the provided test sets, as well as our extended test sets to evaluate it. 
For comparison, we select two typical knowledge graph embedding methods, TransE~\citep{transe_bordes13} and DistMult~\citep{dismult_yang15}, as baselines and compare the performance of MLLMs and MLLMs-A$_{\text{Fusion}}$, enhanced with the knowledge adapter and fusion mechanism (see Appendix~\ref{appendix:implementation} for further implementation details).

\begin{table}[!htbp]
    \centering
    \vspace{-.1cm}
    \caption{Results on the knowledge graph completion task. We attach the number of languages to each type. We can see that for zero-shot languages and unseen languages, using our adapters can significantly improve the performance of LMs on knowledge graph completion.}
    \vspace{-.2cm}
    \resizebox{1.\columnwidth}{!}{
        \begin{tabular}{c | cc | cc | cc}
        \toprule
        \multirow{2}{*}{\bf Model}
        & \multicolumn{2}{c}{\bf Sup.~(5)} 
        & \multicolumn{2}{c}{\bf ZS-In~(15)}
        & \multicolumn{2}{c}{\bf ZS-Un.~(2')} \\
        \cmidrule(lr){2-3}\cmidrule(lr){4-5}\cmidrule(lr){6-7}
        & Hit@1$^\uparrow$ & MRR$^\uparrow$ & Hit@1$^\uparrow$ & MRR$^\uparrow$ & Hit@1$^\uparrow$ & MRR$^\uparrow$ \\
        \midrule
        TransE & \textbf{14.5} & \textbf{23.7} & -/- & -/- & -/- & -/- \\
        DistMult & 8.1 & 14.3 & -/- & -/- & -/- & -/- \\
        \hline
        \hline
        mBERT & 11.2 & 13.8 & 12.8 & 15.7 & 48.2 & 49.1 \\
        mBERT-A$_{\text{Fusion}}$ & 13.1 & 15.7 & \textbf{16.1} & \textbf{18.8} & \textbf{51.8} & \textbf{52.4} \\
        \hline
        XLMR$_{\text{base}}$ & 5.9 & 7.8 & 6.7 & 9.1 & 8.2 & 11.8 \\
        XLMR$_{\text{base}}$-A$_{\text{Fusion}}$ & 9.1 & 11.8 & 10.6 & 13.5 & 16.6 & 19.6 \\
        \hline
        XLMR$_{\text{large}}$ & 7.3 & 9.7 & 8.9 & 11.5 & 16.8 & 20.8 \\
        XLMR$_{\text{large}}$-A$_{\text{Fusion}}$ & 13.1 & 15.6 & 14.3 & 17.3 & 23.9 & 27.4 \\
        \bottomrule
        \end{tabular}
    }
    \vspace{-.5cm}
    \label{tab:mKG-completion}
\end{table}

\paragraph{Results.}
Results are summarized in Table~\ref{tab:mKG-completion} (with further detail in Table~\ref{tab:mKG-completion-detail} of Appendix~\ref{appendix:kg_task}). We report both Hit$@1$ score and Mean Reciprocal Rank (MRR) for evaluation. 
We find that enhancing MLLMs with adapters can improve performance for the \textit{supervised} languages, which is comparable to existing knowledge graph embedding methods. For the \textit{zero-shot} languages and \textit{unseen} languages, existing (transductive) knowledge graph embedding methods cannot perform the task since entities must be in the training set. Here we find that MLLMs still perform comparably to the supervised languages\footnote{Note that due the variable size of entity sets, the task difficulty varies across  languages (see Table~\ref{tab:mKG-completion-detail} in Appendix~\ref{appendix:kg_task}).}, and the enhanced MLLMs-A$_{\text{Fusion}}$ models outperform MLLMs on zero-shot languages by significant margins.
This indicates that the adapters allow factual knowledge to be transferred across languages.

\subsubsection{Entity Alignment}
The entity alignment task is to align entities in different languages. Specifically, given a target language and an entity in a source language (typically English), the model should retrieve that entity from the set of all entities in the target language.

\paragraph{Settings.}
We follow settings of WK3L.\footnote{Even if side information such as the entity description is provided, we only consider the alignment data for simplicity.} Specifically, we train models using the entity alignments English to German, and English to French. We test models on those two \textit{supervised} languages, as well as our extended $17$ \textit{zero-shot} languages and $2$ \textit{unseen} languages.\footnote{Note that we select the extended language only by the size of test set. The ZS-In set is slightly different from the DBP5L, where \textit{ar} and \textit{cs} are newly added and \textit{el} is not included.}
We select one typical MLKG embedding method, MTransE~\citep{mtranse_chen17}, and a state-of-the-art method, JEANS~\citep{chen-etal-2021-cross}, as baselines (see Appendix~\ref{appendix:implementation} for details).

\begin{table}[!htbp]
    \centering
    \vspace{-.1cm}
    \caption{
    Results on multilingual entity alignment tasks. We can find that using our adapters can significantly enhance MLLMs' performance on entity alignment tasks, which also outperforms existing MLKG embedding baselines.}
    \vspace{-.2cm}
    \resizebox{1.\columnwidth}{!}{
        \begin{tabular}{c | cc | cc | cc}
        \toprule
        \multirow{2}{*}{\bf Model}
        & \multicolumn{2}{c}{\bf Sup.~(2)} 
        & \multicolumn{2}{c}{\bf ZS-In (18)}
        & \multicolumn{2}{c}{\bf ZS-Un (2')} \\
        \cmidrule(lr){2-3}\cmidrule(lr){4-5}\cmidrule(lr){6-7}
        & Hit@1 & MRR & Hit@1 & MRR & Hit@1 & MRR \\
        \midrule
        MTransE & 8.7 & 12.5 & -/- & -/- & -/- & -/- \\
        JEANS & 40.0 & 47.5 & -/- & -/- & -/- & -/- \\
        \hline
        \hline
        mBERT & 83.6 & 83.2 & 31.8 & 32.2 & 50.5 & 50.8 \\
        mBERT-A$_{\text{Fusion}}$ & 88.9 & 88.4 & 77.6 & 76.2 & \textbf{91.7} & \textbf{89.3} \\
        \hline
        XLMR$_{\text{base}}$ & 54.8 & 54.8 & 9.4 & 9.7 & 10.9 & 11.1 \\
        XLMR$_{\text{base}}$-A$_{\text{Fusion}}$ & 88.6 & 88.0 & 82.4 & 81.8 & 85.4 & 84.3 \\
        \hline
        XLMR$_{\text{large}}$ & 65.0 & 65.1 & 23.9 & 24.1 & 28.9 & 28.9 \\
        XLMR$_{\text{large}}$-A$_{\text{Fusion}}$ & \textbf{90.2} & \textbf{89.5} & \textbf{90.8} & \textbf{89.0} & 89.8 & 88.5 \\
        \bottomrule
        \end{tabular}
    }
    \vspace{-.5cm}
    \label{tab:mKG-alignment}
\end{table}
\paragraph{Results.}
The results are summarized in Table~\ref{tab:mKG-alignment} (with further detail in Table~\ref{tab:mKG-alignment-detail} of Appendix~\ref{appendix:kg_task}).
Performance is evaluated again by Hits$@1$ and MRR. 
As previously, the (transductive) baselines cannot be extended to languages not in the training set. 
For the \textit{supervised} languages, we can find that existing MLLMs often outperform classic baselines. However, performance of MLLMs on \textit{zero-shot} languages is noticeably worse. This indicates that existing MLLMs do not transfer entity alignment knowledge well to other languages.
However, MLLMs enhanced with the adapter set, MLLMs-A$_{\text{Fusion}}$, generally achieve the best performance, often with significant improvement.
The results indicate that our adapter set successfully enhances MLLMs with multilingual knowledge.

\subsection{MLLM Benchmarks}\label{exp:mllm}
Above results show that our adapter set can enhance MLLMs to perform well on MLKG-related tasks on both previously seen and unseen languages. 
Here, we show that \textit{our knowledge adapter set can allow MLKGs to enhance MLLM performance on language tasks}. In particular, the enhanced MLLMs achieve improved performance on knowledge-intensive language task while maintaining performance on other general language tasks.

\subsubsection{Cross-Lingual Relation Classification}
We select a popular relation classification benchmark: RELX~\citep{koksal-ozgur-2020-relx}, for which MLLMs must extract relations from sentences in a cross-lingual setting. Models are finetuned on a high-resource corpus, and tested on low-resource languages in a zero-shot setting. For this task, MLLMs are required to transfer the knowledge across languages, as well as capture factual knowledge for the relation classification.

\paragraph{Settings.}
Our training data is only in English, and test data contains $4$ more (zero-shot) languages. We follow the exact setting of~\citet{koksal-ozgur-2020-relx} and use the same provided set of hyperparameters to evaluate all MLLMs.
We also report the performance of the enhanced BERT model of~\citet{koksal-ozgur-2020-relx} called Matching the Multilingual Blanks (MTMB) as a baseline.

\begin{table}[!htbp]
    \fontsize{10}{11}\selectfont
    \centering
    \vspace{-.1cm}
    \caption{Results on the multilingual relation classification task (F1 score). We can find that our adapters can effectively enhance MLLMs on the knowledge-intensive downstream tasks, especially for the performance on  zero-shot languages.}
	\vspace{-.2cm}
    \resizebox{.85\columnwidth}{!}{
        \begin{tabular}{c|c|c|c}
        \toprule
        \bf Model
        & \bf Sup.~(En)
        & \bf ZS-In (4)
        & \bf Ave. \\
        \midrule
        mBERT & 61.8 & 57.4 & 58.3 \\
        MTMB & 63.6 & 59.6 & 60.4 \\
        mBERT-A$_{\text{Fusion}}$ & \textbf{64.0} & \textbf{60.9} & \textbf{61.5} \\
        \hline
        XLMR$_{\text{base}}$ & \textbf{61.4} & 56.1 & 57.1 \\
        XLMR$_{\text{base}}$-A$_{\text{Fusion}}$ & 61.3 & \textbf{58.0} & \textbf{58.6} \\
        \hline
        XLMR$_{\text{large}}$ & 63.1 & 59.1 & 59.9 \\
        XLMR$_{\text{large}}$-A$_{\text{Fusion}}$ & \textbf{64.2} & \textbf{60.4} & \textbf{61.1} \\
        \bottomrule
        \end{tabular}
    }
    \vspace{-.5cm}
    \label{tab:relation-classification}
\end{table}
\paragraph{Results.}
Results are summarized in Table~\ref{tab:relation-classification} (see Table~\ref{tab:relation-classification-detail} of Appendix~\ref{appendix:lm_task} for further detail).
We find that for \textit{supervised} languages, mBERT-A$_{\text{Fusion}}$ outperforms both the base version of mBERT as well as the knowledge-enhanced version (MTMB), whereas XLMR with adapters obtains comparable performance.
As for\textit{ zero-shot }languages, MLLMs-A$_{\text{Fusion}}$ achieve consistent and significant improvements over baselines.
This demonstrates that our knowledge adapter set can enhance MLLMs for knowledge-intensive tasks.


\subsubsection{General Language Tasks}
Besides above knowledge-intensive tasks, we show that our knowledge adapter set can maintain the performance of MLLMs on general multilingual language tasks. We select the popular multilingual benchmark called XTREME~\citep{xtreme_hu20} to evaluate the enhanced MLLMs, which are finetuned on English training data, and tested with many other languages. We select cross-lingual NER and QA as two general tasks.
We follow the settings of the XTREME benchmark.  

\begin{table}[!htbp]
    \fontsize{10}{11}\selectfont
    \centering
    \vspace{-.1cm}
    \caption{Results on the multilingual NER task (F1 score). We can find that our adapters can enhance MLLMs on the performance of NER task for zero-shot languages.}
    \vspace{-.2cm}
    \resizebox{.85\columnwidth}{!}{
        \begin{tabular}{c|c|c|c}
        \toprule
        \bf Model
        & \bf Sup.~(En)
        & \bf ZS-In~(39)
        & \bf Ave. \\
        \midrule
        mBERT & \textbf{85.2} & 61.6 & 62.2 \\
        mBERT-A$_{\text{Fusion}}$ & 84.0 & \textbf{62.3} & \textbf{62.9} \\
        \hline
        XLMR$_{\text{large}}$ & 84.7 & 64.9 & 65.4 \\
        XLMR$_{\text{large}}$-A$_{\text{Fusion}}$ & \textbf{85.0} & \textbf{65.3} & \textbf{65.8} \\
        \bottomrule
        \end{tabular}
    }
    \vspace{-.5cm}
    \label{tab:ner}
\end{table}
\paragraph{NER.} 
We select the WikiAnn dataset~\citep{pan-etal-2017-cross} (under the setting of XTREME) for the NER task, where $40$ languages are included for evaluation.   
The results are summarized in Table~\ref{tab:ner}, and detailed results can be found in Table~\ref{tab:ner-detail} in Appendix~\ref{appendix:lm_task}.
We find that MLLMs with our adapter set perform as well as the baseline MLLMs with slight improvements on the zero-shot languages.

\begin{table}[!htbp]
    \centering
    \vspace{-.1cm}
    \caption{Results on the multilingual QA tasks. Using our adapters would not reduce the performance on language modeling tasks, while marginal improvement can even achieved.}
    \vspace{-.2cm}
    \resizebox{.95\columnwidth}{!}{
        \begin{tabular}{c | cc | cc | cc}
        \toprule
        \multirow{2}{*}{\bf Model}
        & \multicolumn{2}{c}{\bf Sup.~(En)} 
        & \multicolumn{2}{c}{\bf ZS-In~(10)}
        & \multicolumn{2}{c}{\bf Ave.} \\
        \cmidrule(lr){2-3}\cmidrule(lr){4-5}\cmidrule(lr){6-7}
        & F1 & EM & F1 & EM & F1 & EM \\
        \midrule
        mBERT & 83.5 & 72.2 & \textbf{62.6} & 47.2 & \textbf{64.5} & 49.4 \\
        mBERT-A$_{\text{Fusion}}$ & \textbf{83.5} & 72.0 & 62.1 & \textbf{47.2} & 62.2 & \textbf{49.5} \\
        \hline
        XLMR$_{\text{large}}$ & 86.5 & 75.7 & 75.6 & 59.3 & 76.6 & 60.8 \\
        XLMR$_{\text{large}}$-A$_{\text{Fusion}}$ & \textbf{88.0} & \textbf{77.6} & \textbf{75.7} & \textbf{59.7} & \textbf{76.8} & \textbf{61.3} \\
        \bottomrule
        \end{tabular}
    }
    \vspace{-.5cm}
    \label{tab:qa}
\end{table}
\paragraph{Question Answering.} 
Following the setting of XTREME, We finetune the models on the SQuAD~\citep{rajpurkar-etal-2016-squad} dataset (in English), and evaluate on the test sets of XQuAD~\citep{artetxe-etal-2020-cross} involving $11$ languges. 
Detailed results are in Table~\ref{tab:qa-detail} in Appendix~\ref{appendix:lm_task}. We find that mBERT-A$_{\text{Fusion}}$ maintains the performance as its original version, while XLMR$_{\text{large}}$-A$_{\text{Fusion}}$ can be boosted slightly. 
In general, MLLMs-A$_{\text{Fusion}}$ with our adapters can obtain comparable or slightly better performance across different language tasks.
For those tasks requiring rich knowledge about triples and entity alignments, our adapter set can indeed enhance the MLLMs.


\subsection{Comparison with Existing Methods}\label{exp:compare}
We compare our approach with the only existing related work~\citep{mlkidamo_liu21} that attempts to integrate MLKGs into MLLMs. However, it only considers a relatively small set of 10 languages and finetunes the entire MLLM with a joint objective, which is computationally expensive. In contrast, as shown below, our knowledge adapter set can achieve better performance at a much lower cost.

\paragraph{Settings.} 
We follow settings and metrics in~\citet{mlkidamo_liu21}, which are slightly different from original settings of RELX and WikiAnn (XTREME) datasets.
We only report the performance for MLLMs that are implemented in their study.


\begin{table}[!htbp]
    \centering
    \vspace{-.1cm}
    \caption{Comparison with~\newcite{mlkidamo_liu21} (denoted by $^\triangle$) on RELX, WikiAnn and XQuAD datasets involving $4$, $10$ and $11$ languages, respectively. We can find that our light adapter-based knowledge enhancement method significantly outperforms previous finetuning-based enhancement method.}
    \vspace{-.2cm}
    \resizebox{1\columnwidth}{!}{
        \begin{tabular}{c | c | c | cc}
        \toprule
        \multirow{2}{*}{\bf Model}
        & \multicolumn{1}{c}{\bf RELX~(4)} 
        & \multicolumn{1}{c}{\bf WikiAnn~(10)}
        & \multicolumn{2}{c}{\bf XQuAD~(11)} \\
        \cmidrule(lr){2-2}\cmidrule(lr){3-3}\cmidrule(lr){4-5}
        & Acc. & F1 & F1 & EM \\
        \midrule
        mBERT & 60.1 & -/- & -/- & -/- \\
        mBERT$^\triangle$ & 61.1 & -/- & -/- & -/- \\
        mBERT-MLKG & \textbf{64.7} & -/- & -/- & -/- \\
        \hline
        XLMR$_{\text{base}}$ & 56.7 & -/- & -/- & -/- \\
        XLMR$_{\text{base}}^\triangle$ & 58.3 & -/- & -/- & -/- \\
        XLMR$_{\text{base}}$-A$_{\text{Fusion}}$ & \textbf{61.7} & -/- & -/- & -/- \\
        \hline
        XLMR$_{\text{large}}$ & 61.3 & \textbf{68.5} & 76.6 & 60.8 \\
        XLMR$_{\text{large}}^\triangle$ & 61.9 & 66.9 & 76.5 & 60.6 \\
        XLMR$_{\text{large}}$-A$_{\text{Fusion}}$ & \textbf{64.6} & 67.6 & \textbf{76.8} & \textbf{61.3} \\
        \bottomrule
        \end{tabular}
    }
    \vspace{-.5cm}
    \label{tab:damo}
\end{table}

\begin{figure*}[!htbp]
	\centering
	\subfigure{
	    \pgfplotsset{width=6cm, height=4.cm}
		\begin{tikzpicture}[font=\tiny]
		\begin{axis}[ybar, bar width=0.16cm, enlarge x limits=.1, legend columns=-1,
                    symbolic x coords={mBERT, mBERT-A$_{\text{EP}}$, mBERT-A$_{\text{TP}}$, mBERT-A$_{\text{ES}}$, mBERT-A$_{\text{TS}}$, mBERT-A$_{\text{Large}}$, mBERT-A$_{\text{Fusion}}$}, 
                    y label style={at={(axis description cs:0.2,0.5)},anchor=south},
                    x tick label style={rotate=45, anchor=east, align=left},
                    ylabel={MLKG Completion Performance}, 
                    nodes near coords, 
                    nodes near coords align={vertical},
                    every node near coord/.append style={font=\fontsize{1}{1}\selectfont},
                    ymin=10., ymax=26, xtick={mBERT, mBERT-A$_{\text{EP}}$, mBERT-A$_{\text{TP}}$, mBERT-A$_{\text{ES}}$, mBERT-A$_{\text{TS}}$, mBERT-A$_{\text{Large}}$, mBERT-A$_{\text{Fusion}}$}]
		\addplot coordinates {(mBERT, 14.1) (mBERT-A$_{\text{EP}}$, 11.0) (mBERT-A$_{\text{TP}}$, 16.6) (mBERT-A$_{\text{ES}}$, 13.4) (mBERT-A$_{\text{TS}}$, 13.9) (mBERT-A$_{\text{Large}}$, 15.2) (mBERT-A$_{\text{Fusion}}$, 17.1)};
		\addlegendentry{Hit$@1$}
		\addplot coordinates {(mBERT, 16.8) (mBERT-A$_{\text{EP}}$, 14.1) (mBERT-A$_{\text{TP}}$, 19.1) (mBERT-A$_{\text{ES}}$, 16.2) (mBERT-A$_{\text{TS}}$, 16.6) (mBERT-A$_{\text{Large}}$, 18.1) (mBERT-A$_{\text{Fusion}}$, 19.7)};
		\addlegendentry{MRR}
		\end{axis}
		\end{tikzpicture}
	}
	\subfigure{
	    \pgfplotsset{width=6cm, height=4.cm}
		\begin{tikzpicture}[font=\tiny]
		\begin{axis}[ybar, bar width=0.16cm, enlarge x limits=0.1, legend columns=-1,
                    symbolic x coords={mBERT, mBERT-A$_{\text{EP}}$, mBERT-A$_{\text{TP}}$, mBERT-A$_{\text{ES}}$, mBERT-A$_{\text{TS}}$, mBERT-A$_{\text{Large}}$, mBERT-A$_{\text{Fusion}}$}, 
                    y label style={at={(axis description cs:0.2,0.5)},anchor=south},
                    x tick label style={rotate=45, anchor=east, align=left},
                    ylabel={Entity Alignment Performance}, 
                    nodes near coords, 
                    nodes near coords align={vertical},
                    every node near coord/.append style={font=\fontsize{1}{1}\selectfont},
                    ymin=20., ymax=115, xtick={mBERT, mBERT-A$_{\text{EP}}$, mBERT-A$_{\text{TP}}$, mBERT-A$_{\text{ES}}$, mBERT-A$_{\text{TS}}$, mBERT-A$_{\text{Large}}$, mBERT-A$_{\text{Fusion}}$}]
		\addplot coordinates {(mBERT, 38) (mBERT-A$_{\text{EP}}$, 86) (mBERT-A$_{\text{TP}}$, 27) (mBERT-A$_{\text{ES}}$, 77) (mBERT-A$_{\text{TS}}$, 47) (mBERT-A$_{\text{Large}}$, 68) (mBERT-A$_{\text{Fusion}}$, 79)};
		\addlegendentry{Hit$@1$}
		\addplot coordinates {(mBERT, 38) (mBERT-A$_{\text{EP}}$, 84) (mBERT-A$_{\text{TP}}$, 28) (mBERT-A$_{\text{ES}}$, 76) (mBERT-A$_{\text{TS}}$, 48) (mBERT-A$_{\text{Large}}$, 67) (mBERT-A$_{\text{Fusion}}$, 78.0)};
		\addlegendentry{MRR}
		\end{axis}
		\end{tikzpicture}
	}
    \subfigure{
	    \pgfplotsset{width=4.5cm, height=4.cm}
		\begin{tikzpicture}[font=\tiny]
		\begin{axis}[ybar, bar width=0.25cm, enlarge x limits=.1, legend columns=-1,
                    symbolic x coords={mBERT, mBERT-A$_{\text{EP}}$, mBERT-A$_{\text{TP}}$, mBERT-A$_{\text{ES}}$, mBERT-A$_{\text{TS}}$, mBERT-A$_{\text{Large}}$, mBERT-A$_{\text{Fusion}}$}, 
                    y label style={at={(axis description cs:0.25,0.5)},anchor=south},
                    x tick label style={rotate=45, anchor=east, align=left},
                    ylabel={RELX Performance: F1}, 
                    nodes near coords, 
                    nodes near coords align={vertical},
                    ymin=57., ymax=62.5, xtick={mBERT, mBERT-A$_{\text{EP}}$, mBERT-A$_{\text{TP}}$, mBERT-A$_{\text{ES}}$, mBERT-A$_{\text{TS}}$, mBERT-A$_{\text{Large}}$, mBERT-A$_{\text{Fusion}}$}]
		\addplot coordinates {(mBERT, 58.3) (mBERT-A$_{\text{EP}}$, 59.2) (mBERT-A$_{\text{TP}}$, 57.6) (mBERT-A$_{\text{ES}}$, 60.9) (mBERT-A$_{\text{TS}}$, 61.4) (mBERT-A$_{\text{Large}}$, 59.1) (mBERT-A$_{\text{Fusion}}$, 61.5)};
		\end{axis}
		\end{tikzpicture}
	}
	\vspace{-.7cm}
	\caption{Ablation study results. We select two MLKG-related tasks and the relation classification task for evaluation. We can find that adapters that integrate factual knowledge into MLLMs achieve better performance than others on the MLKG completion task, while adapters that integrate cross-lingual alignments outperform others on the entity alignment task. For the relation classification task, sentence-level adapters achieve better performance. For our adapter set, it can achieve roughly the best performance under all conditions.}
	\vspace{-.5cm}
	\label{fig:ablation}
\end{figure*}
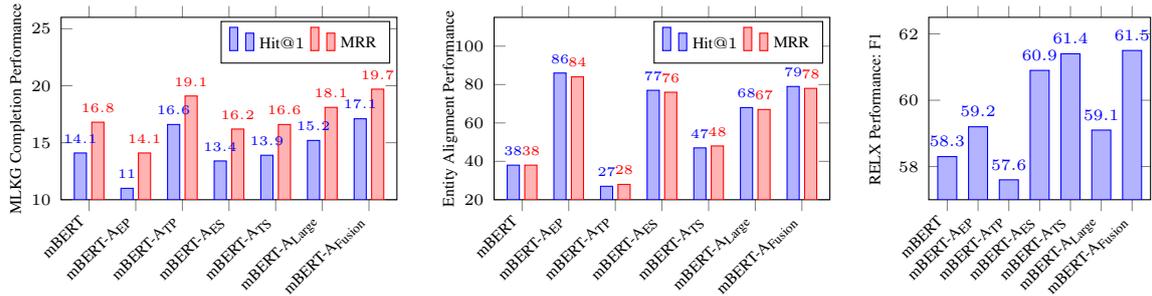

\paragraph{Results.} In Table~\ref{tab:damo}, for the relation classification task, where \citet{mlkidamo_liu21} outperforms the MLLM baseline, our method achieves significant further improvement.
%
For NER, only $10$ popular \textit{zero-shot} languages (instead of $40$ languages in XTREME) are selected for their knowledge integration and evaluation. Even if generally our method achieves better performance for XMLR$_{\text{large}}$-A$_{\text{Fusion}}$ ($40$ languages) in Table~\ref{tab:ner}, it performs slightly worse than the original version here ($10$ popular languages). However, the performance of~\citet{mlkidamo_liu21} is worse still.
For QA, similar performance is achieved by all three MLLMs, although our enhanced MLLM slightly outperforms other methods.

\subsection{Ablation Study}\label{exp:ablation}
We conduct ablation studies to understand our knowledge adapters and show that they work as expected.\footnote{Since the improvements brought by our adapters are consistent across different MLLMs, we mainly consider mBERT for analysis. We report results on the knowledge graph completion, entity alignment and relation classification tasks, which each require different aspects of knowledge.}
We also compare against a large adapter (A$_\text{Large}$) with a comparable total number of parameters (including fusion layers). The large adapter is trained with the same settings as our adapter set and has one set of parameters that integrate all knowledge types at once.
As previously, we finetune the original mBERT, mBERT-A$_\text{Large}$, and mBERT with our adapters on each downstream task.

In Figure~\ref{fig:ablation}, for the knowledge graph completion task (\textit{left}), mBERT-A$_\text{TP}$ and mBERT-A$_\text{TS}$ perform better than their entity-based counterparts. While mBERT-A$_\text{Large}$ also performs well, mBERT-A$_\text{Fusion}$ outperforms it significantly. 
For the entity alignment task (\textit{center}), the situation is reversed such that better performance is achieved by mBERT-A$_\text{EP}$ and, mBERT-A$_\text{ES}$. Our mBERT-A$_\text{Fusion}$ also achieves comparable performance which is much better than mBERT-A$_\text{Large}$ with shared parameters.
As for the relation classification task (\textit{right}), sentence-level adapters outperform phrase-level adapters, which is intuitive since the task requires sentence-level context. Fusing all four adapters (i.e., mBERT-A$_\text{Fusion}$) gives the best performance while mBERT-A$_\text{Large}$ performs worse than single smaller adapters.
In summary, with our method,  we learn different types of knowledge in separate adapters, which can be fused in different proportions according to the downstream task at hand to typically perform better and more consistently than any single adapter-enhanced MLLMs.


%% file: main_text/5conclusion.tex
\section{Other Related Work}
\paragraph{MLLM for MLKG.}
Several works use the implicit knowledge in language models to improve knowledge graph-related tasks~\citep{kgbert_yao19,cakekgc_niu22}. However, these approaches are for monolingual knowledge triples and can not easily incorporate cross-lingual entity alignment. \citet{mlkgc_gnn_huang22} use MLLMs for knowledge graph completion, but language models only encode entities, and the task itself is achieved by graph neural networks.
Previous MLKG embedding methods consider entity alignment~\citep{mtranse_chen17,chen-etal-2020-multilingual}, but are designed for existing MLKGs, and can not generalize to other, e.g.\ low-resource, languages without the multilingual knowledge in MLLMs~\citep{pires-etal-2019-multilingual,wu-dredze-2019-beto}. 

\paragraph{MLKG for MLLM.}
\citet{mlkidamo_liu21} propose to synthesize \textit{code-switched} sentences to solve the problem but the resulting MLKG-enhanced MLLMs achieve minimal improvement on language understanding tasks as shown in our experiment, and it cannot benefit the MLKG field.
%
In summary, our work first combine MLKG and MLLM, showing that combining them using our light knowledge adapter set can effectively improve the downstream task performance on both sides. 


\section{Conclusion}
In this paper we propose an approach to enhance MLLMs with MLKGs using a set of knowledge adapters, where explicit knowledge from MLKGs is integrated into the implicit knowledge learned by MLLMs. In experiments, we show that enhanced MLLMs can conduct MLKG-related tasks and achieve better performance on knowledge-intensive tasks, especially on low-resource languages where knowledge graphs are not available.

\section*{Limitations}
We point out that there are some limitations of our work. First, even if the adapter set can enhance MLLMs to perform well on various downstream tasks, it is not suitable for tasks with the fully zero-shot setting (without any training data), since the fusion module has to be tuned to suit the task. Second, as shown in our results, the fusion module cannot always outperform all single adapters. For some tasks, a better fusion mechanism could be proposed for the improvement.

\section*{Reproducibility Statement}
We elaborate the experiment settings and hyperparameters in the paper and in Appendix~\ref{appendix:implementation}. We have published our prepossessed multilingual knowledge integration data, extended MLKG-related task datasets, as well as our code.

\section*{Ethics Statement}
We do not foresee any significant ethical concerns in this work.

\section*{Acknowledgments}
We are grateful to the anonymous reviewers for their insightful comments and suggestions.
Yifan Hou is supported by the Swiss Data Science Center PhD Grant (P22-05). 
Carl Allen is supported by an ETH AI Centre Postdoctoral Fellowship.
We also acknowledge support from an ETH Zurich Research grant (ETH-19 21-1) and a grant from the Swiss National Science Foundation (project \#201009) for this work.


%% file: main_text/6appendix.tex
\section{Implementation Details}\label{appendix:implementation}
We implement the adapters using the AdapterHub library\footnote{\href{https://adapterhub.ml/}{https://adapterhub.ml/}}, where all Transformer layers in MLLMs are inserted with adapters. 

\paragraph{Adapters in Knowledge Enhancement.} 
To train these knowledgeable adapters, we use $8$ GPUs (Tesla V100) with batch size as $128$. The learning rate is set as $1e-4$. We use the Adam optimizer with $1e4$ warm-up steps. We train Adapter-EP by randomly sampling entity alignments in different languages. The number of sampled alignments is around $94.2$ million. And the training epoch number for Adapter-TP, Adapter-ES, and Adapter-TS is all set as $10$. As for the InfoNCE loss, we use the negative sampling within batch. Since we train adapters with sampling strategy and use the contrastive learning loss instead of mask language modeling, it only takes several hours to train one adapter (1-10 hours). The whole enhancement procedure would take around half a day.

\paragraph{Adapters in knowledge graph completion.} For MLLM-based methods, we set all hyperparameters as the same to ensure the comparison is fair\footnote{Note that users may search more fine-grained hyperparameters, but the relative performance would not change.}. We use the average value of word(-piece) representation as the entity embedding. Specifically, we train MLLMs as well as MLLMs-AF (including adapters and the fusion mechanism) to embed entities, where the output representations of the object entities should be close to the context (subject entities with relations) output representations. The similarity is measure by cosine\footnote{We also tried different metrics but find that cosine distance works well in this task.}. During the training, the learning rate is set as $1e-8$, and the epoch number is set as $10$. The batch size is set as $8$. We train MLLMs using the contrastive learning loss similar to the knowledge integration process. 

\paragraph{Adapters in Entity Alignment.} Similarly, we set all hyperparameters as the same for all MLLM-based methods. Specifically, we set the epoch number as $1$ since the overfitting is easy with training data only on $2$ languages. Other hyperparameters and settings are the same to that of the MLKG Completion task.

\paragraph{Adapters in Language Tasks.} We evaluate our adapter set with MLLMs on the XTREME benchmark. The evaluation settings are the same as theirs.

\section{Knowledge Integration Dataset Statistics}\label{appendix:ki_dataset}
The detailed statistics can be found in Table~\ref{tab:distribution-wikidata} below.
\begin{table*}[!htbp]
    \centering
    \caption{Distribution of Wikidata for adapter training. We report the full name and ISO code for all languages. For the entity, relation, and triple, we report the ratio of the label in that specific language to the total number of it.}
	\resizebox{2\columnwidth}{!}{
        \begin{tabular}{ccccc || ccccc || ccccc }
        \toprule
        ISO & Lang. & Entity (\%) & Relation (\%) & Triple (\%) & ISO & Lang. & Entity (\%) & Relation (\%) & Triple (\%) & ISO & Lang. & Entity (\%) & Relation (\%) & Triple (\%)  \\
        \midrule
        af  & Afrikaans & 56.4  & 20.5 & 31.8 & gu  & Gujarati & 12.2 & 14.2 & 2.1  & nn  & Norwegian Nynorsk & 70.6 & 44.4 & 57.9 \\
        an  & Aragonese & 59.8  & 0.7  & 10.7 & he  & Hebrew & 25.3 & 62.2 & 29.7 & no  & Norwegian & 0.0  & -    & -   \\
        ar  & Arabic & 33.5  & 91.0 & 42.3 & hi  & Hindi & 14.8 & 13.3 & 4.4  & oc  & Occitan & 60.9 & 23.9 & 32.1 \\
        ast & Asturian & 84.2  & 28.3 & 71.3 & hr  & Croatian & 57.8 & 23.1 & 33.5 & pl  & Polish & 92.5 & 73.0 & 85.9 \\
        az  & Azerbaijani & 19.3  & 19.3 & 9.8  & hu  & Hungarian & 71.9 & 64.6 & 70.2 & pt  & Portuguese & 96.4 & 80.8 & 91.0 \\
        bar & Bavarian & 52.4  & 1.8  & 10.0 & hy  & Armenian & 21.5 & 21.4 & 17.0 & ro  & Romanian & 81.7 & 32.8 & 59.3 \\
        be  & Belarusian & 18.0  & 52.7 & 11.6 & id  & Indonesian & 65.2 & 48.2 & 47.7 & ru  & Russian & 54.4 & 88.6 & 64.1 \\
        bg  & Bulgarian & 31.9  & 22.6 & 19.4 & is  & Icelandic & 52.3 & 7.6  & 15.0 & scn & Sicilian & 39.6 & 25.5 & 17.7 \\
        bn  & Bengali & 18.3  & 34.6 & 11.5 & it  & Italian & 97.8 & 78.2 & 97.0 & sco & Scots & 56.8 & 27.3 & 27.3 \\
        br  & Breton & 54.5  & 18.8 & 30.0 & ja  & Japanese & 37.5 & 77.5 & 48.3 & sh  & Serbo-Croatian & 21.7 & 9.2  & 6.9  \\
        bs  & Bosnian & 44.7  & 27.3 & 18.6 & jv  & Javanese & 41.3 & 1.6  & 7.1  & sk  & Slovak & 62.4 & 25.8 & 38.4 \\
        ca  & Catalan & 87.2  & 99.3 & 88.9 & ka  & Georgian & 16.2 & 23.8 & 9.3  & sl  & Slovenian & 69.1 & 24.8 & 56.0 \\
        ceb & Cebuano & 51.5  & 0.3  & 0.2  & kk  & Kazakh & 16.7 & 4.0  & 2.2  & sq  & Albanian & 73.0 & 28.1 & 47.2 \\
        cs  & Czech & 73.5  & 68.4 & 66.4 & kn  & Kannada & 13.7 & 7.8  & 2.1  & sr  & Serbian & 23.3 & 92.6 & 17.5 \\
        cy  & Welsh & 61.4  & 35.4 & 43.3 & ko  & Korean & 26.2 & 58.2 & 25.3 & sv  & Swedish & 91.7 & 73.8 & 90.9 \\
        da  & Danish & 77.3  & 57.5 & 75.4 & la  & Latin & 59.9 & 9.4  & 23.1 & sw  & Swahili & 50.4 & 0.6  & 6.2  \\
        de  & German & 98.5  & 90.7 & 98.5 & lb  & Luxembourgish & 55.3 & 25.2 & 33.5 & ta  & Tamil & 14.8 & 18.8 & 4.8  \\
        el  & Greek & 19.5  & 46.5 & 16.5 & lt  & Lithuanian & 52.5 & 15.7 & 27.4 & te  & Telugu & 13.2 & 17.7 & 3.1  \\
        en  & English & 100.0 & 100.0& 100.0& lv  & Latvian & 38.2 & 40.2 & 25.6 & th  & Thai & 16.2 & 20.5 & 7.3  \\
        es  & Spanish & 98.7  & 94.0 & 98.6 & mk  & Macedonian & 16.5 & 95.1 & 9.3  & tl  & Tagalog & 16.4 & 7.2  & 5.3  \\
        et  & Estonian & 60.8  & 25.9 & 40.9 & ml  & Malayalam & 15.9 & 14.6 & 4.7  & tr  & Turkish & 64.4 & 81.2 & 50.9 \\
        eu  & Basque & 74.0  & 37.4 & 54.4 & mn  & Mongolian & 11.5 & 1.3  & 0.2  & tt  & Tatar & 19.0 & 35.7 & 12.4 \\
        fa  & Persian & 32.5  & 51.1 & 33.7 & mr  & Marathi & 13.4 & 17.4 & 3.7  & uk  & Ukrainian & 45.2 & 97.7 & 44.4 \\
        fi  & Finnish & 89.9  & 56.3 & 78.8 & ms  & Malay & 56.3 & 40.9 & 35.0 & ur  & Urdu & 16.7 & 28.1 & 7.8  \\
        fr  & French & 98.5  & 97.3 & 99.1 & my  & Burmese & 11.7 & 5.3  & 0.9  & uz  & Uzbek & 17.1 & 3.7  & 4.6  \\
        fy  & Western Frisian & 41.6  & 4.7  & 7.6  & nds & Low Saxon & 54.1 & 23.1 & 29.3 & vi  & Vietnamese & 74.7 & 32.8 & 44.4 \\
        ga  & Irish & 78.4  & 25.2 & 57.3 & ne  & Nepali & 11.3 & 7.7  & 1.1  & war & Waray-Waray & 61.1 & 0.1  & 0.0  \\
        gl  & Galician & 65.2  & 38.5 & 45.9 & nl  & Dutch & 98.3 & 100.0& 98.2 & zh  & Chinese & 41.1 & 64.9 & 49.7 \\
        \bottomrule
        \end{tabular}
    }
    \label{tab:distribution-wikidata}
\end{table*}

\section{MLKG Dataset Statistics and Detailed Results}\label{appendix:kg_task}
The detailed statistics and results can be found in Table~\ref{tab:mKG-completion-detail} and Table~\ref{tab:mKG-alignment-detail}.
\begin{table*}[!htbp]
	\caption{The performance of various models for the MLKG completion task (Hit@1/MRR) across different languages. We also report the number of entities in the test set to show the general difficulty of the completion task in that language.}
	\vspace{-.3cm}
	\smallskip
	\centering
	\resizebox{2\columnwidth}{!}{
		\smallskip\begin{tabular}{c|c|cc|cc|cc|cc}
		    \toprule
            Language & \# of test set & TransE & DisMult & mBERT & mBERT-MLKG & XLM & XLM-MLKG & XLM-R & XLM-R-MLKG \\
			\midrule
			el & 1082 & 13.1 / 24.3 & 8.9 / 9.8 & 9.2 / 11.6 & 8.5 / 11.2 & 4.8 / 6.9 & 6.9 / 9.7 & 5.0 / 7.5 & 9.3 / 12.8 \\
            en & 5984 & 7.3 / 16.9 & 8.8 / 18.3 & 15.2 / 17.7 & 18.5 / 21.3 & 8.2 / 10.0 & 11.7 / 14.8 & 10.4 / 12.5 & 17.5 / 19.9 \\
            es & 4101 & 13.5 / 24.4 & 7.4 / 13.2 & 14.3 / 17.2 & 17.7 / 20.5 & 7.0 / 9.4 & 11.7 / 14.9 & 9.7 / 12.2 & 18.0 / 20.7 \\
            fr & 4436 & 17.5 / 27.6 & 6.1 / 14.5 & 12.7 / 15.4 & 17.4 / 19.9 & 6.3 / 8.5 & 12.1 / 14.5 & 9.2 / 11.6 & 16.2 / 18.5 \\
            ja & 2569 & 21.1 / 25.3 & 9.3 / 15.8 & 4.6 / 6.9 & 3.6 / 5.8 & 3.1 / 4.4 & 3.0 / 5.0 & 2.4 / 4.6 & 4.7 / 7.4 \\
            \hline
            ast & 2823 & - & - & 13.9 / 16.8 & 19.1 / 21.8 & 7.1 / 9.5 & 13.7 / 16.5 & 10.6 / 12.9 & 17.5 / 20.5 \\
            ca & 2959 & - & - & 14.8 / 17.6 & 19.1 / 21.5 & 7.9 / 10.4 & 13.8 / 16.5 & 11.1 / 13.4 & 17.4 / 20.2 \\
            da & 2566 & - & - & 16.1 / 19.2 & 19.9 / 23.0 & 8.7 / 11.6 & 13.3 / 16.9 & 11.5 / 14.1 & 17.6 / 21.4 \\
            de & 4059 & - & - & 14.1 / 16.8 & 17.4 / 20.4 & 8.3 / 11.2 & 11.4 / 14.6 & 9.8 / 12.6 & 15.6 / 18.7 \\
            fa & 2329 & - & - & 5.0 / 7.1 & 5.3 / 6.9 & 3.9 / 4.8 & 4.1 / 5.8 & 5.1 / 7.3 & 5.2 / 7.2 \\
            fi & 2582 & - & - & 11.2 / 14.6 & 16.1 / 19.1 & 6.2 / 8.6 & 9.9 / 13.0 & 8.2 / 11.1 & 13.7 / 17.0 \\
            hu & 2558 & - & - & 13.7 / 16.7 & 18.4 / 21.4 & 6.4 / 9.2 & 11.4 / 14.8 & 10.0 / 12.5 & 15.7 / 18.7 \\
            it & 3614 & - & - & 14.4 / 17.0 & 17.3 / 19.8 & 7.6 / 9.8 & 12.2 / 15.2 & 10.4 / 12.8 & 15.7 / 18.6 \\
            nb & 2717 & - & - & 16.4 / 19.4 & 19.5 / 23.3 & 8.9 / 11.6 & 13.5 / 17.0 & 11.3 / 13.9 & 18.0 / 21.4 \\
            nl & 4316 & - & - & 14.0 / 16.8 & 19.1 / 21.7 & 7.3 / 9.8 & 13.3 / 15.9 & 8.6 / 11.5 & 17.4 / 20.2 \\
            pl & 2998 & - & - & 13.4 / 17.2 & 18.6 / 21.8 & 6.1 / 8.5 & 9.7 / 13.3 & 8.7 / 11.5 & 14.6 / 18.0 \\
            pt & 3184 & - & - & 15.4 / 18.4 & 18.0 / 20.6 & 7.3 / 9.7 & 12.3 / 15.4 & 9.6 / 12.1 & 17.5 / 20.6 \\
            ru & 2887 & - & - & 9.4 / 11.8 & 10.3 / 12.1 & 3.5 / 5.5 & 4.6 / 6.6 & 4.8 / 7.4 & 6.3 / 8.6 \\
            sv & 2993 & - & - & 15.7 / 18.5 & 18.7 / 22.0 & 9.2 / 11.7 & 13.0 / 16.4 & 11.0 / 13.6 & 17.8 / 21.3 \\
            zh & 2591 & - & - & 5.1 / 7.4 & 4.1 / 6.4 & 2.2 / 4.2 & 2.7 / 5.1 & 3.4 / 5.3 & 4.3 / 6.8 \\
            \hline
            eo & 963 & - & - & - & - & 8.2 / 11.8 & 16.6 / 19.6 & 16.8 / 20.8 & 23.9 / 27.4  \\
            vo & 164 & - & - & 48.1 / 49.1 & 51.8 / 52.4 & - & - & - & -  \\
			\bottomrule
		\end{tabular}
	}	
	\label{tab:mKG-completion-detail}
	\vspace{-.2cm}
\end{table*} 
\begin{table*}[!htbp]
	\caption{The performance of various models for the entity alignment task (Hit@1/MRR) across different languages. We also report the number of entities in the test set to show the general difficulty of the completion task in that language.}
	\vspace{-.3cm}
	\smallskip
	\centering
	\resizebox{2\columnwidth}{!}{
		\smallskip\begin{tabular}{c|c|cc|cc|cc|cc}
		\toprule
            Language & \# of test set & MTransE & JEANS & mBERT & mBERT-MLKG & XLM & XLM-MLKG & XLM-R & XLM-R-MLKG \\
			\midrule
			en->fr & 39155 & 14.0 / 17.7 & 46.3 / 53.8 & 87.1 / 86.4 & 92.6 / 92.1 & 55.3 / 55.3 & 92.1 / 91.4 & 65.2 / 65.2 & 93.5 / 92.8 \\
            en->de & 41018 & 3.4 / 7.2 & 33.7 / 41.2 & 80.1 / 79.9 & 85.2 / 84.7 & 54.3 / 54.3 & 85.1 / 84.6 & 64.8 / 64.9 & 86.8 / 86.2 \\
            \hline
            en->ar & 16818 & - & - & 8.9 / 10.0 & 68.6 / 67.4 & 0.7 / 0.9 & 63.4 / 62.5 & 0.9 / 1.1 & 81.8 / 80.0 \\
            en->ast & 19834 & - & - & 41.8 / 41.9 & 85.2 / 83.9 & 13.0 / 13.2 & 93.6 / 92.6 & 33.5 / 33.8 & 97.3 / 96.3 \\
            en->ca & 22567 & - & - & 38.2 / 38.3 & 81.5 / 80.2 & 10.8 / 11.1 & 90.2 / 88.8 & 29.9 / 30.2 & 94.5 / 93.4 \\
            en->cs & 16570 & - & - & 40.0 / 40.3 & 82.5 / 81.1 & 11.9 / 12.2 & 89.8 / 88.6 & 30.4 / 30.5 & 93.9 / 92.8 \\
            en->da & 20093 & - & - & 39.2 / 39.4 & 82.4 / 81.3 & 12.7 / 12.9 & 91.7 / 90.5 & 33.0 / 33.2 & 95.5 / 94.4 \\
            en->es & 28288 & - & - & 40.6 / 40.3 & 81.8 / 80.2 & 11.5 / 11.7 & 90.1 / 88.6 & 33.2 / 32.3 & 94.3 / 92.7 \\
            en->fa & 16120 & - & - & 10.1 / 11.3 & 69.4 / 68.2 & 1.0 / 1.2 & 67.6 / 66.9 & 1.8 / 2.2 & 83.1 / 81.8 \\
            en->fi & 20608 & - & - & 39.4 / 39.4 & 81.3 / 79.9 & 12.4 / 12.6 & 90.0 / 88.8 & 32.2 / 32.4 & 94.2 / 93.1 \\
            en->hu & 18896 & - & - & 36.3 / 36.7 & 80.5 / 79.4 & 11.3 / 11.4 & 89.2 / 88.0 & 29.6 / 29.9 & 93.6 / 92.7 \\
            en->it & 26393 & - & - & 39.4 / 39.5 & 80.2 / 78.7 & 11.5 / 11.8 & 88.4 / 86.9 & 31.2 / 31.2 & 92.4 / 91.0 \\
            en->ja & 22012 & - & - & 8.9 / 10.1 & 64.3 / 63.4 & 0.7 / 0.8 & 60.9 / 60.0 & 1.4 / 1.5 & 77.8 / 76.4 \\
            en->nb & 20748 & - & - & 39.2 / 39.3 & 82.5 / 81.1 & 11.5 / 11.8 & 91.8 / 90.4 & 32.2 / 32.5 & 95.6 / 94.4 \\
            en->nl & 29378 & - & - & 41.3 / 41.3 & 82.4 / 80.5 & 12.2 / 12.4 & 90.8 / 89.0 & 34.1 / 34.1 & 94.6 / 92.8 \\
            en->pl & 21535 & - & - & 38.7 / 38.9 & 80.0 / 78.7 & 11.2 / 11.4 & 87.6 / 86.3 & 30.2 / 30.3 & 92.6 / 91.2 \\
            en->pt & 23001 & - & - & 41.5 / 41.5 & 82.5 / 81.0 & 12.3 / 12.6 & 90.6 / 89.2 & 33.1 / 33.2 & 94.4 / 93.1 \\
            en->ru & 22665 & - & - & 19.2 / 20.2 & 74.2 / 72.6 & 3.6 / 3.7 & 78.0 / 76.2 & 10.0 / 10.3 & 87.9 / 85.9 \\
            en->sv & 22986 & - & - & 39.7 / 39.7 & 81.6 / 80.1 & 11.8 / 12.1 & 90.6 / 89.2 & 32.2 / 32.4 & 94.4 / 93.1 \\
            en->zh & 20891 & - & - & 10.2 / 11.1 & 55.1 / 54.8 & 9.0 / 10.6 & 49.8 / 49.5 & 1.9 / 1.9 & 67.3 / 66.2 \\
            \hline
            en->eo & 8913 & - & - & - & - & 10.9 / 11.1 & 85.4 / 84.3 & 28.9 / 28.9 & 89.8 / 88.5  \\
            en->vo & 2954 & - & - & 50.5 / 50.8 & 91.7 / 89.3 & - & - & - & -  \\
			\bottomrule
		\end{tabular}
	}	
	\label{tab:mKG-alignment-detail}
	\vspace{-.2cm}
\end{table*}

\section{MLLM Dataset Statistics and Detailed Results}\label{appendix:lm_task}
The detailed statistics and results can be found in Table~\ref{tab:relation-classification-detail} (relation classification), Table~\ref{tab:ner-detail} (name entity recognition), and Table~\ref{tab:qa-detail} (question answering).
\begin{table*}[!htbp]
	\caption{Detailed results of the cross-lingual relation classification task (RELX) evaluated by F1 score.}
	\vspace{-.3cm}
	\smallskip
	\centering
		\smallskip\begin{tabular}{c|cc|cc|cc}
		\toprule
            Language & mBERT & mBERT-MLKG & XLM & XLM-MLKG & XLM-R & XLM-R-MLKG \\
			\midrule
            {en} & 61.8 & 64.0 & 61.4 & 61.3 & 63.1 & 64.2  \\
            \hline
            {de} & 57.5 & 60.0 & 57.5 & 56.1 & 58.0 & 60.2  \\
            {es} & 57.9 & 63.1 & 56.9 & 59.7 & 59.8 & 60.7  \\
            {fr} & 58.3 & 61.1 & 55.7 & 58.0 & 59.5 & 61.5  \\
            {tr} & 55.8 & 59.3 & 54.1 & 58.0 & 59.1 & 59.0  \\
            \hline
            {average} & 58.3 & 61.5 & 57.1 & 58.6 & 59.9 & 61.1  \\
			\bottomrule
		\end{tabular}
	\label{tab:relation-classification-detail}
	\vspace{-.2cm}
\end{table*} 

\begin{table*}[!htbp]
	\caption{Detailed results of the NER task (Wikiann) evaluated by F1 socre.}
	\vspace{-.3cm}
	\smallskip
	\centering
	\resizebox{2\columnwidth}{!}{
		\smallskip\begin{tabular}{c|cc|cc||c|cc|cc}
		\toprule
            Language & mBERT & mBERT-MLKG & XLM-R & XLM-R-MLKG & Language & mBERT & mBERT-MLKG & XLM-R & XLM-R-MLKG \\
			\midrule
            \textbf{en} & 85.2 & 84.0 & 84.7 & 85.0 & {ka} & 64.6 & 66.9 & 71.6 & 69.3  \\
            {af} & 77.4 & 77.2 & 78.9 & 79.2 & {kk} & 45.8 & 49.1 & 56.2 & 53.3  \\
            {ar} & 41.1 & 40.5 & 53.0 & 51.8 & {ko} & 59.6 & 60.2 & 60.0 & 61.0  \\
            {bg} & 77.0 & 76.2 & 81.4 & 80.6 & {ml} & 52.3 & 53.1 & 67.8 & 61.0  \\
            {bn} & 70.0 & 72.8 & 78.8 & 78.1 & {mr} & 58.2 & 55.0 & 68.1 & 67.2  \\
            {de} & 78.0 & 78.6 & 78.8 & 78.5 & {ms} & 72.7 & 68.1 & 57.1 & 74.6  \\
            {el} & 72.5 & 70.8 & 79.5 & 79.6 & {my} & 45.2 & 55.5 & 54.3 & 56.8  \\
            {es} & 77.4 & 74.8 & 79.6 & 75.8 & {nl} & 81.8 & 82.3 & 84.0 & 83.5  \\
            {et} & 75.4 & 78.6 & 79.1 & 78.1 & {pt} & 80.8 & 78.7 & 81.9 & 82.5  \\
            {eu} & 66.3 & 68.3 & 60.9 & 59.0 & {ru} & 64.0 & 66.8 & 69.1 & 70.5  \\
            {fa} & 46.2 & 38.6 & 61.9 & 48.9 & {sw} & 67.5 & 70.1 & 70.5 & 70.0  \\
            {fi} & 77.2 & 78.3 & 79.2 & 79.0 & {ta} & 50.7 & 53.8 & 59.5 & 60.8  \\
            {fr} & 79.6 & 78.9 & 80.5 & 80.2 & {te} & 48.5 & 48.2 & 55.8 & 50.9  \\
            {he} & 56.6 & 54.4 & 56.8 & 57.9 & {th} & 3.6 & 0.1 & 1.3 & 2.9  \\
            {hi} & 65.0 & 66.3 & 73.0 & 73.0 & {tl} & 71.7 & 74.6 & 73.2 & 78.0  \\
            {hu} & 76.4 & 78.0 & 79.8 & 80.6 & {tr} & 71.8 & 74.4 & 76.1 & 80.6  \\
            {id} & 53.5 & 54.6 & 53.0 & 55.9 & {ur} & 36.9 & 43.9 & 56.4 & 63.2  \\
            {it} & 81.5 & 81.6 & 81.3 & 81.2 & {vi} & 71.8 & 70.7 & 79.4 & 78.9  \\
            {ja} & 29.0 & 29.2 & 23.2 & 23.1 & {yo} & 44.9 & 50.9 & 33.6 & 45.4  \\
            {jv} & 66.4 & 65.3 & 62.5 & 66.9 & {zh} & 42.7 & 45.0 & 33.1 & 28.9  \\
			\bottomrule
		\end{tabular}
	}	
	\label{tab:ner-detail}
	\vspace{-.2cm}
\end{table*}

\begin{table*}[!htbp]
	\caption{Detailed results of the QA task (XQuAD) evaluated by F1/EM score.}
	\vspace{-.3cm}
	\smallskip
	\centering
		\smallskip\begin{tabular}{c|cc|cc}
		\toprule
            Language & mBERT & mBERT-MLKG & XLM-R & XLM-R-MLKG \\
			\midrule
            {en} & 83.5 / 72.2 & 83.5 / 72.0 & 86.5 / 75.7 & 88.0 / 77.6  \\
            {ar} & 61.5 / 45.1 & 61.3 / 44.5 & 68.6 / 49.0 & 76.2 / 58.9  \\
            {de} & 70.6 / 54.0 & 70.6 / 54.8 & 80.4 / 63.4 & 79.6 / 62.8  \\
            {el} & 62.6 / 44.9 & 63.5 / 47.5 & 79.8 / 61.7 & 79.1 / 61.3  \\
            {es} & 75.5 / 56.9 & 74.4 / 57.2 & 82.0 / 63.9 & 82.4 / 64.4  \\
            {hi} & 59.2 / 46.0 & 57.2 / 42.9 & 76.7 / 59.7 & 75.6 / 59.3  \\
            {ru} & 71.3 / 53.3 & 70.5 / 54.4 & 80.1 / 64.3 & 79.7 / 63.6  \\
            {th} & 42.7 / 33.5 & 43.6 / 36.8 & 74.2 / 62.8 & 73.3 / 61.2  \\
            {tr} & 55.4 / 40.1 & 53.7 / 38.0 & 75.9 / 59.3 & 74.9 / 58.9  \\
            {vi} & 69.5 / 49.6 & 67.7 / 47.9 & 79.1 / 59.0 & 80.0 / 60.6  \\
            {zh} & 58.0 / 48.3 & 58.0 / 48.3 & 59.3 / 50.0 & 56.0 / 46.7  \\
            \hline
            average & 64.5 / 49.4 & 62.2 / 49.5 & 76.6 / 60.8 & 76.8 / 61.3  \\
			\bottomrule
		\end{tabular}
	\label{tab:qa-detail}
	\vspace{-.2cm}
\end{table*}